\documentclass{article}

\usepackage[preprint]{neurips_2022}

\usepackage[utf8]{inputenc} % allow utf-8 input
\usepackage[T1]{fontenc}    % use 8-bit T1 fonts
\usepackage{hyperref}       % hyperlinks
\usepackage{url}            % simple URL typesetting
\usepackage{booktabs}       % professional-quality tables
\usepackage{amsfonts}       % blackboard math symbols
\usepackage{nicefrac}       % compact symbols for 1/2, etc.
\usepackage{microtype}      % microtypography
\usepackage{xcolor}         % colors
\usepackage{makecell}

\usepackage{makecell}
\usepackage{tabularx, multirow}
\usepackage{amsmath}
\usepackage{amssymb}
\usepackage{mathtools}
\usepackage{amsthm}
\usepackage{bm}
\usepackage{bbold}
\usepackage{acro}           % acronyms
\usepackage{svg}

\theoremstyle{plain}
\newtheorem{theorem}{Theorem}[section]
\newtheorem{proposition}[theorem]{Proposition}
\newtheorem{lemma}[theorem]{Lemma}
\newtheorem{corollary}[theorem]{Corollary}
\theoremstyle{definition}

\theoremstyle{remark}
\newtheorem{remark}[theorem]{Remark}

\DeclareAcronym{ift} {
  short = IFT,
  long = Implicit Function Theorem
}

\DeclareAcronym{snn} {
  short = SNN,
  long = Spiking Neural Network
}

\DeclareAcronym{lif} {
  short = LIF,
  long = Leaky Integrate-and-Fire
}

\DeclareAcronym{if} {
  short = IF,
  long = Integrate-and-Fire}

\DeclareAcronym{srm} {
  short = SRM,
  long = Spike Response Model
}

\DeclareAcronym{bptt} {
  short = BPTT,
  long = back-propagation through time
}

\def\mindex#1{\index{#1}}

\def\sq{\hbox{\rlap{$\sqcap$}$\sqcup$}}
\def\qed{\ifmmode\sq\else{\unskip\nobreak\hfil
\penalty50\hskip1em\null\nobreak\hfil\sq
\parfillskip=0pt\finalhyphendemerits=0\endgraf}\fi\medskip}

\long\def\defbox#1{\framebox[.9\hsize][c]{\parbox{.85\hsize}{%
\parindent=0pt
\baselineskip=12pt plus .1pt      % STYLE
\parskip=6pt plus 1.5pt minus 1pt % CHANGES
 #1}}}

\long\def\beginbox#1\endbox{\subsection*{}%
\hbox{\hspace{.05\hsize}\defbox{\medskip#1\bigskip}}%
\subsection*{}}

\def\endbox{}

\newsavebox{\junk}
\savebox{\junk}[1.6mm]{\hbox{$|\!|\!|$}}

\def\det{{\mathop{\rm det}}}

\def\bR{{\mathbb R}}

\def\bfs{{\bf s}}

\def\bfz{{\bf z}}

\def\ttx{{\mathtt x}}
\def\tty{{\mathtt y}}

\def\bfmath#1{{\mathchoice{\mbox{\boldmath$#1$}}%
{\mbox{\boldmath$#1$}}%
{\mbox{\boldmath$\scriptstyle#1$}}%
{\mbox{\boldmath$\scriptscriptstyle#1$}}}}

\def\bfmY{\bfmath{Y}}

\def\bfmhhaY{\bfmath{\hhaY}} %\widehat{\widehat{Y}}}}
\def\bfmhhaY{\hbox to 0pt{$\widehat{\bfmY}$\hss}\widehat{\phantom{\raise 1.25pt\hbox{$\bfmY$}}}}

\def\til={{\widetilde =}}

\def\clL{{\cal L}}

 \def\FRAC#1#2#3{\genfrac{}{}{}{#1}{#2}{#3}}

\def\ddtp{{\mathchoice{\FRAC{1}{d^{\hbox to 2pt{\rm\tiny +\hss}}}{dt}}%
{\FRAC{1}{d^{\hbox to 2pt{\rm\tiny +\hss}}}{dt}}%
{\FRAC{3}{d^{\hbox to 2pt{\rm\tiny +\hss}}}{dt}}%
{\FRAC{3}{d^{\hbox to 2pt{\rm\tiny +\hss}}}{dt}}}}

\def\average#1,#2,{{1\over #2} \sum_{#1}^{#2}}

\def\eye(#1){{\bf(#1)}\quad}

\newcounter{rmnum}

\newcounter{anum}

{\end{list}}

\def\ass(#1:#2){(#1\ref{#1:#2})}

\def\ritem#1{
\item[{\sf \ass(\current_model:#1)}]
}

\newenvironment{recall-ass}[1]{%
\begin{description}
\def\current_model{#1}}{
\end{description}
}

\long\def\comment#1{}

\newfont{\bbb}{msbm10 scaled 700}

\newfont{\bb}{msbm10 scaled 1100}

\renewcommand{\det}{{\hbox{det}}}

\def\ul{\mathbf{u}^{(l)}}
\def\zl{\mathbf{z}^{(l)}}
\def\al{\mathbf{a}^{(l)}}
\def\sigl{\bm{\sigma}^{(l)}}
\def\Jdl{\mathbf{J}^{\mathcal{D}^{(l)}}}
\def\Jil{\mathbf{J}^{\mathcal{I}^{(l)}}}

\def\all{\mathbf{a}^{(l-1)}}
\def\zlL{\mathbf{z}^{(l+1)}}
\def\Wl{\mathbf{W}^{(l)}}
\def\wl{\mathbf{w}^{(l)}}

\def\el{\mathbf{e}^{(l)}}
\def\dl{\bm{d}^{(l)}}

\def\loss{\mathcal{L}}
\def\boldl#1{\bm{#1}^{(l)}}
\def\boldx#1#2{\bm{#1}^{(#2)}}
\def\deriv#1#2{\frac{\partial #1}{\partial #2}}

\title{EXODUS: Stable and Efficient Training of Spiking Neural Networks}

\author{%
  Felix C. Bauer \\
  \And
  Gregor Lenz \\
  \And
  Saeid Haghighatshoar \\
  \And
  Sadique Sheik \\
  \and
  SynSense AG \\
  8050 Zurich, Switzerland \\
  \texttt{name.surname@synsense.ai} \\
}

\begin{document}

\maketitle

\begin{abstract}
\acp{snn} are gaining significant traction in machine learning tasks where energy-efficiency is of utmost importance. Training such networks using the state-of-the-art \acf{bptt} is, however, very time-consuming.
Previous work by~\citet{shrestha_orchard18} employs an efficient GPU-accelerated back-propagation algorithm called SLAYER, which speeds up training considerably. SLAYER, however, does not take into account the neuron reset mechanism while computing the gradients, which we argue to be the source of numerical instability. To counteract this, SLAYER introduces a gradient scale hyper parameter across layers, which needs manual tuning. 
In this paper, 
(i)\,we modify SLAYER and design an algorithm called EXODUS, that accounts for the neuron reset mechanism and applies the \acf{ift} to calculate the correct gradients (equivalent to those computed by BPTT),
(ii)\,we eliminate the need for ad-hoc scaling of gradients, thus, reducing the training complexity tremendously,
(iii)\,we demonstrate, via computer simulations, that EXODUS is numerically stable and achieves a comparable or better performance than SLAYER especially in various tasks with \acp{snn} that rely on temporal features.
Our code is available at https://github.com/synsense/sinabs-exodus.
\end{abstract}

\section{Introduction}\label{intro}
Spiking Neural Networks (SNNs) are a class of biologically-inspired networks with single bit activations, fine-grained temporal resolution and highly sparse outputs. Their memory makes them especially suitable for sequence tasks and their sparse output promises extremely low power consumption, especially when combined with an event-based sensor and asynchronous hardware~\citep{goltz2021fast, davies2021advancing}. SNNs have thus garnered considerable attention for machine learning tasks that aim to achieve low power consumption and/or biological realism~\citep{diehl_unsupervised_snn_2015, cao_snn_object_recognition_2015, roy_snn_computing_2019, panda_snn_residual_2020, comsa_temporal_coding_2020}.

%Spiking Neural Networks (SNNs) are a new class of artificial neural networks (ANNs) that have recently gained significant traction in machine learning~\citep{maass_third_gen}. In contrast with  conventional ANNs, which consist of memory-less neurons, thus, perform only spatial processing across their layers, SNNs are able to process their signals both spatially across their layers and temporally within each individual layer. This feature gives SNNs a tremendous flexibility in signal processing tasks with an intrinsic temporal nature such as speech recognition  (e.g., keyword spotting). 

%The computation model in SNNs is inspired by biological neuronal mechanisms where the communication between neurons is done via temporally-sparse spike signals. This allows to implement SNNs as asynchronous circuits, which consume 1-2 orders of magnitude less power than the conventional clock-based synchronous chips (Von-Neumann computation structure)~\citep{true_north_2015, loihi_2018}. Due to this feature, SNNs have garnered considerable attention recently~\citep{diehl_unsupervised_snn_2015, cao_snn_object_recognition_2015, roy_snn_computing_2019, panda_snn_residual_2020, comsa_temporal_coding_2020} as low-power alternatives for machine learning tasks. 

SNNs are notoriously difficult to train unfortunately due to the highly nonlinear neuron dynamics, extreme output quantization and potential internal state resets, even for relatively simple neuron models such as Integrate-and-Fire (IF) or Leaky-Integrate-and-Fire (LIF)~\citep{gerstner95, burkitt_integrate_fire_2006}). 
Different approaches and learning rules have thus emerged to train SNNs.   
%A number of works has tried to address the difficulty of training SNNs and c
% include methods for time-based learning such as SpikeProp, Fast and Deep
Biologically-inspired local learning rules~\citep{hebbian_overview_2013, lee_stdp_pretrain_2018, lobov_stdp_2020} do not need to rely on a global error signal, but often fail to scale to larger architectures~\citep{bartunov2018assessing}. 
%Methods that convert pre-trained ANNs to SNNs~\citep{rueckauer_conversion_2017, Ding2021OptimalAC, ho_conversion_2021}, 
Methods that work directly with spike timings to calculate gradients bypass the issue of non-differentiable spikes, but are typically limited to time-to-first spike encoding~\citep{bohte2000spikeprop, goltz2021fast, mostafa2017supervised}.
Fueled by the success of deep learning, surrogate gradient methods have more recently paved the way for flexible gradient-based optimization in SNNs with the aim to close the accuracy gap to ANN counterparts~\citep{huh_gd_NEURIPS2018, neftci_etal19, safa_convsnn_2021}.

Surrogate gradient methods make use of a smoothed output activation (usually a function of internal state variables) in the neuron during the backward pass to approximate the discontinuous activation. This is well-supported in modern deep learning frameworks and allows the direct application of \acf{bptt} to train SNNs. This alone would make it feasible to train SNNs successfully were it not for the high temporal resolution needed in SNNs. Activation is typically very sparse across time and when processing data from an event-based sensor, time resolution can decrease to micro-seconds. This makes it necessary to simulate input using a lot of time steps on today's von Neumann machines, which work in discrete time. \ac{bptt} with $T$ discrete-time steps incurs a computational complexity of order $O(T^2)$ and a memory overhead of $O(T)$ on such architectures, which makes training SNNs tremendously slow and expensive.

To alleviate this issue,~\citet{shrestha_orchard18} proposed SLAYER, an algorithm in which gradients are back-propagated across layers as usual, but they are computed jointly in time such that the $O(T)$ complexity due to \emph{sequential} back-propagation in time is fairly eliminated by highly parallelized and GPU-accelerated 
%(C++\footnote{https://bitbucket.org/bamsumit/slayer} and Python\footnote{https://github.com/bamsumit/slayerPytorch} implementations, both under GNU GPLv3) 
joint gradient computation. 
Mathematically speaking, SLAYER can be seen as gradient computation for a modified forward computation graph underlying the chain rule, in which all the time dynamics is contracted into a single node. However, this comes at the cost of creating a loop in the computation graph where the gradients cannot be back-propagated via chain rule. To solve this issue, SLAYER ignores a loopy term, known as reset kernel, which models the effect of output spike generation on the internal neuron potential. 
%This approximation seems to be fine in a dense spike regime (over-triggered state) but may create a large inconsistency  when the spikes are very sparse (under-triggered state).
As a result, SLAYER yields a considerable speed-up for training SNNs at the cost of some numerical instability.
SLAYER typically deals with this instability by tweaking a hyper parameter which scales the gradient magnitude. This needs considerable hand-tuning and scales unfavorably to deeper architectures and longer time sequences.
In this paper we propose EXODUS (EXact computation Of Derivatives as Update to SLAYER), in which we address these numerical issues and compute gradients equivalently to what BPTT computes, while at the same time achieving a significant speedup of 1-2 orders of magnitude. In summary, our contributions are as follows:

(i)\,
We improve the SLAYER algorithm by taking into account the reset response of neurons to their output firing times.

(ii)\,
By applying the implicit function theorem, we resolve the loopy structure in each layer's computation graph and are able to compute correct gradients that can be back-propagated through each layer. %We call our proposed algorithm EXact calculation Of Derivatives as Update to Slayer (EXODUS).

(iii)\,
We eliminate the need for ad-hoc scaling of gradients, needed for solving the numerical instability of SLAYER, thus, reducing the training complexity tremendously.

(iv)\,
We demonstrate, via numerical simulations, that EXODUS is robust to changes in gradient scaling and achieves comparable or better performance than SLAYER in various tasks with \acp{snn} that rely on temporal features.

\section{Preliminaries and background}

\subsection{Implicit Function Theorem}
\label{ss:ift}
In many problems in statistics, mathematics, control theory, machine learning, etc. the  state of a problem is represented in terms of a collection of variables. However, in many cases, these variables are correlated due to existence of constraints. Here, we are interested in a setting where one may have a collection of $m+n$ variables and a set of $m$ equations (equality constraints) connecting them together. Since there are $m$ equations, one may hope to solve for $m$ variables, at least locally, as a function of the remaining $n$ variables. It is conventional to call the first $m$ variables \textit{dependent} and the remaining $n$ variables \textit{independent} as the latter may vary (at least locally) independently of one another while the values of the former depend on the specific choice of those $n$ independent variables.

The Implicit Function Theorem (IFT) provides rigorous conditions under which this is possible and specifies when the $m$ dependent variables are differentiable with respect to $n$ independent ones.

\begin{theorem}[Implicit Function Theorem]
Let $\phi: \mathbb{R}^n \times \mathbb{R}^m \to \mathbb{R}^m$ be a differentiable function, let $\mathcal{Z}=\{(x,y) \in \mathbb{R}^n \times \mathbb{R}^m: \phi(x,y)=0\}$ be the zero-set of $\phi$, and let $(x_0,y_0) \in \mathcal{Z}$ be an arbitrary point in $\mathcal{Z}$. If the $m\times m$ matrix $\frac{\partial \phi}{\partial y}(x_0,y_0)$ is non-singular, i.e., $\det\big ( \frac{\partial \phi}{\partial y} (x_0,y_0) \big ) \not = 0$, then, 
\begin{itemize}
    \item there is an open neighborhood $\mathcal{N}_\ttx$ around $x_0$ and an open neighborhood $\mathcal{N}_\tty$ around $y_0$ such that $\frac{\partial \phi}{\partial y} (x,y)$ is non-singular for all $(x,y) \in \mathcal{N}:=\mathcal{N}_\ttx \times \mathcal{N}_\tty$ (including of course the original point $(x_0,y_0)$). 
    
    \item there is a function $\psi: \mathcal{N}_\ttx \to \mathcal{N}_\tty$ such that $(x,\psi(x))$ belongs to the zero-set $\mathcal{Z}$, i.e., $\phi(x,\psi(x))=0$, for all $x \in \mathcal{N}_\ttx$.
    Therefore, $y=\psi(x)$ can be written as function of $x$.
    %; therefore, the variables $y$ in $\mathcal{N}_\mathrm{y}$ can be written as a function $y=\psi(x)$ of the variables $x$ in $\mathcal{N}_\mathrm{x}$.
    
    \item (chain rule) $\psi$ is a differentiable function of $x$  in $\mathcal{N}_\ttx$ and 
    \begin{align}
        \frac{\partial \phi}{\partial y}  \times \frac{\partial \psi}{\partial x} + \frac{\partial \phi}{\partial x}=0,
        \label{eq:ift_1}
    \end{align}
    which from the non-singularity of $\frac{\partial \phi}{\partial y}$ yields
    \begin{align}
         \frac{\partial \psi}{\partial x} = - \Big ( \frac{\partial \phi}{\partial y} \Big ) ^ {-1} \times \frac{\partial \phi}{\partial x}.
        \label{eq:ift_2}
    \end{align}
    
\end{itemize}

\end{theorem}

\begin{remark}
Note that here, for simplicity, we denoted the independent and dependent variables with $x\in \mathbb{R}^n$ and $y \in \mathbb{R}^m$. In general one may choose any disjoint subsets of the variables of size $m$ and $n$ as dependent and independent variables and verify the conditions of the IFT.
\end{remark}

%Illustrative examples on how to apply the \ac{ift} for finding derivatives of dependent variables with respect to independent variables, and where  intuitive ad-hoc application of the chain rule may fail, can be found in Appendix~\ref{appdx:ift_examples}.

In Appendix~\ref{appdx:ift_examples}, we provide examples to illustrate how \ac{ift} is applied for computing the derivative of the dependent variables with respect to independent ones. In particular, we show that intuitive ad-hoc application of the chain rule in a loopy computation graph and neglecting the conditions of \ac{ift} may indeed yield wrong results. 

\subsection{Spike Response Model}\label{ss:srm}
Neuron dynamics in SNNs can be described by a state-space model where the internal state of each neuron depends on both its current input and its previous states. When applying the conventional back-propagation algorithm, therefore, the gradients need to be back-propagated not only through layers of the network but also through time. 
In this paper, we will focus on SLAYER (Spike LAYer Error Reassignment) training algorithm~\citep{shrestha_orchard18}. SLAYER is based on the spike response model (SRM,~\citep{gerstner95}), where the state of a spiking neuron at each time instant $n$ is described by its membrane potential \(u[n]\), given by
\begin{align}
    u[n] &= \sum_i w_i (\bm{\epsilon} \ast s_i^{in})[n] + (\bm{\nu} \ast s^{out})[n-1],\label{state_space_1}\\
    s^{out}[n] &= f_s(u[n]),\label{state_space_2}.
\end{align}
Here \(s^{in}_i[n]\) and $w_i$ denote the input spikes received from and the weight of the \(i\)-th presynaptic neuron, where $\bm{\epsilon}$ and $\bm{\nu}$ denote the spike response and reset kernel of the neuron (with $\ast$ denoting the convolution operation in time) to the incoming and outgoing spikes,  where a discrete delay of size $1$ is introduced for the reset kernel $\bm{\nu}$ to defer the effect of outgoing spikes to the next time instant. 

%QUESTION: Use binary spike train or allow multiple spikes per time step?
Outgoing spikes are obtained from the membrane potential through a memory-less binary spike generating function \(f_s: \bR \to \{0,1\}\) where $f_s(u[n])=1$ when the membrane potential \(u[n]\) reaches or exceeds the neuron firing threshold \(\theta > 0\), and is \(0\) otherwise. 
Since $f_s$ is not differentiable, a common solution for obtaining well-defined gradients for training SNNs is to use $f_s$ in the forward pass to produce outgoing spikes but replace it with a differentiable function in the backward pass, where the gradients are computed through back-propagation algorithm. This is known as the surrogate gradient method in SNN literature.
Different surrogate gradients have been proposed such as piecewise linear~\citep{esser2016convolutional, bohte2011error}, tanh~\citep{wozniak2020deep}, or exponential functions~\citep{shrestha_orchard18}. In this paper, with some abuse of notation, we denote the surrogate gradient by $f_s'(.)$. Our derivations are valid for any \(f_s'\) as long as $f_s'(u[n])$ is well-defined for all feasible values of the membrane potential \(u[n]\) at all time instants $n$. 

\subsection{Vectorized network model}
In this paper, as in~\citet{shrestha_orchard18}, we focus on a  feedforward network architecture with \(L\) layers. Using \eqref{state_space_1} and \eqref{state_space_2} and applying vectorization, we may write the forward dynamics of a layer \(l\) with \(N_l\) neurons and input weights \(\mathbf{W}^{(l-1)} \in \mathbb{R}^{N_l \times N_{l-1}}\) more compactly (and equivalently) as:
\begin{align}
    \mathbf{a}^{(l-1)}[n] &= (\bm{\epsilon} \ast \mathbf{s}^{(l-1)})[n], \label{eq:a}\\
    \mathbf{z}^{(l)}[n] &= \mathbf{W}^{(l-1)} \mathbf{a}^{(l-1)}[n], \label{eq:z}\\
    \mathbf{u}^{(l)}[n] &= \mathbf{z}^{(l)}[n] + (\bm{\nu} \ast \mathbf{s}^{(l)})[n-1],  \label{eq:u}\\
    \mathbf{s}^{(l)}[n] &= f_s(\mathbf{u}^{(l)}[n]), \label{eq:s}
\end{align}
Here, $\mathbf{a}^{(l-1)}[n]$ is introduced as auxiliary variable for the input spikes (output spikes of previous layer) after being filtered/smoothed out by the neuron spike response $\bm{\epsilon}$ (thus, post-synaptic response), and $\mathbf{z}^{(l)}$ for the weighted version of these post-synaptic signals. 

We describe the model in post-synaptic terms, taking \(\{\mathbf{a}^{(l-1)}[n]: n\in [T]\}\) and \(\{\mathbf{a}^{(l)}[n]: n\in [T]\}\) as the input and output of a specific layer $l\in [L]$ across $T$ time instants, where we used the short-hand notation $[N]=\{0,1,\dots, N-1\}$. The model output is given by \(\{\mathbf{a}^{(L)}[n]: n\in [T]\}\). We also define the loss as \(\mathcal{L}(\mathbf{a}^{L}[0], ..., \mathbf{a}^{L}[T-1])\) in terms of the network output over all time instants $[T]$.

\begin{figure*}[t]
    \centering
    \includegraphics[width=\textwidth]{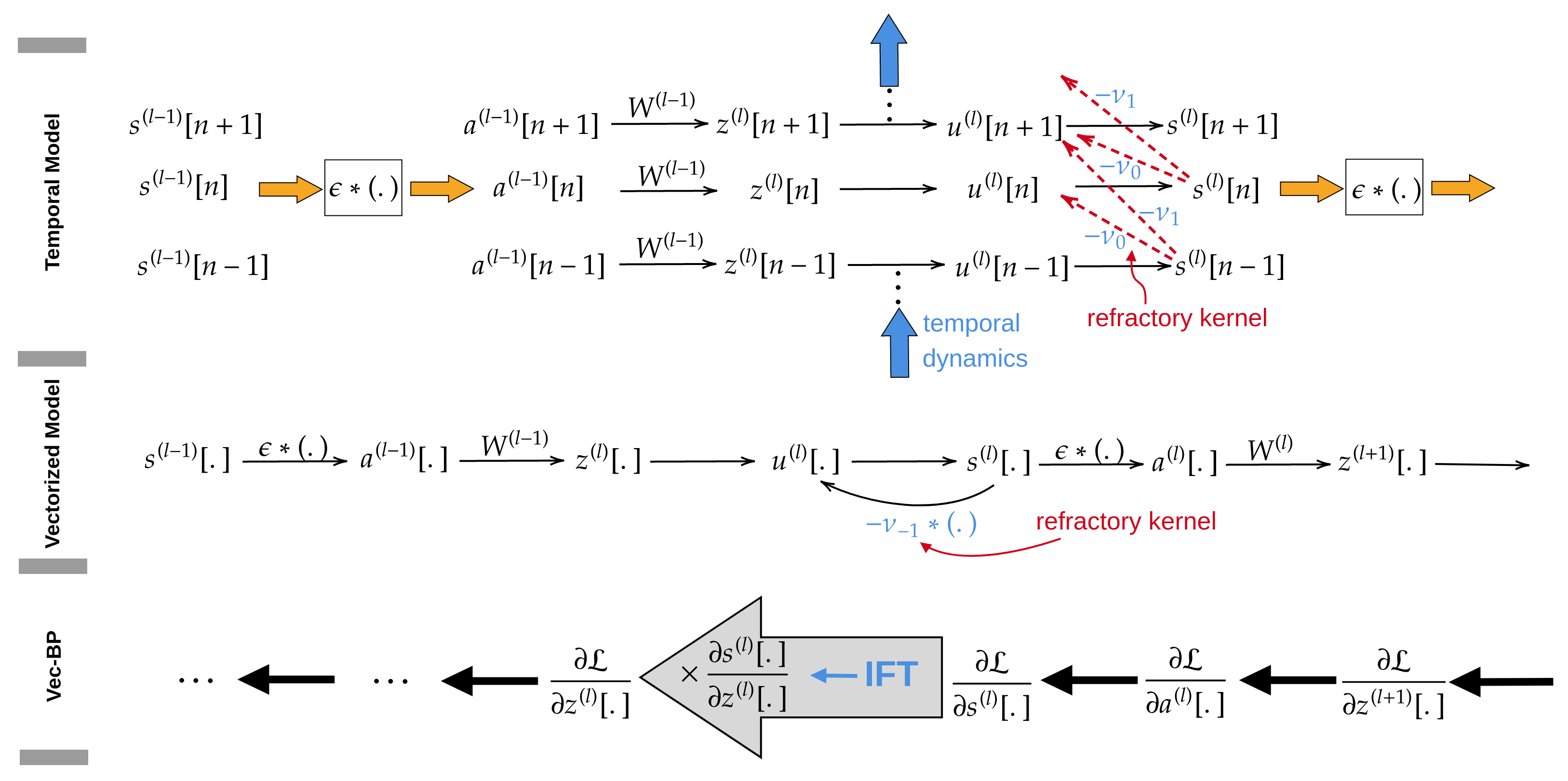}
    \caption{\textbf{Upper:} Directed acyclic computation graph along temporal and spatial dimension. \textbf{Middle:} Loopy computation graph after vectorization along the temporal dimension. The loop is introduced by the reset kernel and is ignored in the backward path of SLAYER. \textbf{Lower:} Vectorized backward path, applying IFT to back-propagate the gradients through the loopy computation graph.}
    \label{fig:vectorized_model}
\end{figure*}

The computational graph of the described network model has a DAG (directed acyclic graph) structure in spatio-temporal dimensions, as illustrated in Fig.\,\ref{fig:vectorized_model}. Therefore, the gradients can be propagated from the loss \(\mathcal{L}\) to the trainable weights backward across the layers (spatial) and also time instants (temporal). However, as explained above, this approach is computationally slow. 

The SLAYER algorithm introduced by~\citet{shrestha_orchard18} avoids this by vectorizing the variables over time, as illustrated in Fig.\,\ref{fig:vectorized_model}. This,  however, introduces loops in the computational graph due to the mutual dependence of the vectorized variables \(\ul[.]\) and \(\boldl{s}[.]\) in equations \eqref{eq:u} and \eqref{eq:s}. This prohibits back-propagating through these variables. 
To solve this issue, SLAYER ignores a term known as reset kernel (effect of output spikes on neuron potential) in the calculation of its gradients. This approximation seems to be fine in a dense spike regime (over-triggered state) but creates  inconsistency  when the spikes are very sparse (under-triggered state).

\section{Derivation of EXODUS Gradients}\label{ss:grads}
\subsection{Exact Vector Back-Propagation}
We calculate gradients precisely by taking into account the reset kernel neglected by SLAYER. To do so, we apply the \ac{ift} to back-propagate the gradients through the loopy computation graph. As the loops occur only between the variables \(\ul[.]\) and \(\boldl{s}[.]\) within the same layer, we apply IFT to each layer $l \in [L]$ individually. More specifically, by applying the chain rule, we back-propagate the gradients from the last layer to compute $\deriv{\clL}{\bfs^{(l)}[.]}$. We show that, although there is a loop between \(\ul[.]\) and \(\boldl{s}[.]\), we are still able to compute $\deriv{\bfs^{(l)}[.]}{\bfz^{(l)}[.]}$ (see, Fig.\,\ref{fig:vectorized_model}). Multiplying this gradient with $\deriv{\clL}{\bfs^{(l)}[.]}$ enables to compute  
$\deriv{\clL}{\bfz^{(l)}[.]}$, which is the gradient back-propagated to the previous layer.

\subsection{Derivations for a generic model}
To apply IFT, we need to specify the underlying equations and also the dependent and independent variables.  
Let us consider \eqref{eq:u} and \eqref{eq:s} for all $n\in[T]$:
\begin{align*}
    \mathbf{\Phi}^{(l)}_{u}[n] &:= \mathbf{u}^{(l)}[n] - \mathbf{z}^{(l)}[n] - (\bm{\nu} \ast \mathbf{s}^{(l)})[n-1] = 0,\\
    \mathbf{\Phi}^{(l)}_{s}[n] &:= \mathbf{s}^{(l)}[n] - f_s(\mathbf{u}^{(l)}[n]) = 0.
\end{align*}
This is a system of \(2 N_l T\) equations in terms of three vectorized variables $\mathbf{u}^{(l)}[.], \mathbf{s}^{(l)}[.],\mathbf{z}^{(l)}[.]$, each of dimension $N_lT$. Therefore, by a simple dimensionality check, we can see that we may write two of these variables (dependent) as a differentiable function of the other variable (independent) provided that the conditions of IFT hold.

To pass the chain rule through the loopy computation graph at layer $l$, we need to compute $\deriv{\mathbf{s}^{(l)}[.]}{\mathbf{z}^{(l)}[.]}$ (see, e.g., Fig.\,\ref{fig:vectorized_model}). This implies that we need to treat $\mathbf{z}^{(l)}[.]$ as the independent and $(\mathbf{s}^{(l)}[.], \mathbf{u}^{(l)}[.])$ as the dependent variables. For simplicity of notation, we denote these independent and dependent variables and the corresponding equations  by
\begin{align*}
    \mathcal{I}^{(l)} :&=\Big \{ \mathbf{z}^{(l)}[n]: n \in [T]\Big \}, \quad
    \mathcal{D}^{(l)} :=\Big \{\mathbf{u}^{(l)}[n], \mathbf{s}^{(l)}[n] : n \in [T]\Big \}, \\
    \boldl{\Phi} :&=\Big \{\mathbf{\Phi}^{(l)}_{u}[n], \mathbf{\Phi}^{(s)}_{u}[n]: n \in [T] \Big \}.
\end{align*}
Let \(\Jdl =\deriv{\boldl{\Phi}}{\mathcal{D}^{(l)}} \in \mathbb{R}^{2N_lT \times N_lT}\) and \(\Jil= \deriv{\boldl{\Phi}}{\mathcal{I}^{(l)}}\in \mathbb{R}^{2N_lT \times N_lT}\) be the Jacobian matrices of the equations \(\boldl{\Phi}\) with respect to the dependent and independent variables,  respectively. 
Let us also define \(\mathbf{G}^{(l)}=\deriv{\mathcal{D}^{(l)}}{\mathcal{I}^{(l)}} \in \mathbb{R}^{2N_lT \times N_lT}\) as the gradients of dependent variables with respect to the independent ones. 

With this, we verify the IFT conditions: (i)\,All the equations are differentiable, provided that \(f_s\) is a differentiable function. (ii) By bringing \(\Jdl\) into a row-echelon form, we can prove that \(\Jdl\) is non-singular. We refer to Appendix~\ref{appdx:sol} for a detailed derivation. The gradients \(\mathbf{G}^{(l)}\) are then found by solving IFT equation \eqref{eq:ift_1} \(\Jdl \cdot \mathbf{G}^{(l)} = -\Jil\).

Applying a forward substitution method (see Appendix~\ref{appdx:sol}) yields the desired gradients:

\begin{align}
    \sigl_n[m]  := \Big ( \deriv{\boldl{s}[.]}{\zl[.]} \Big )_{m, n} =\deriv{\boldl{s}[m]}{\zl[n]} = \begin{cases}
        0 & m<n \\
        \boldl{f'}[m] & m=n \\
        \boldl{f'}[m] \big(\bm{\nu} \ast \sigl_n \big) [m-1] & m>n , \label{eq:dsdz}
    \end{cases}
\end{align}
where \(\boldl{f'}[m]\) is the \emph{diagonal matrix} holding the surrogate gradients \((\boldl{f'}[m])_{ii} = f_s'(u_i^{(l)}[m])\).

As we explained before, computing $\deriv{\boldl{s}[.]}{\zl[.]}$ via IFT allows us to push the  back-propagation (chain rule) through loops in computational graph.
The remaining steps needed for back-propagation are quite straightforward and are obtained from equations \eqref{eq:z} and \eqref{eq:a} as follows:
\begin{align}
    \deriv{\zl[m]}{\all[n]} &= \delta_{m,n} \cdot \boldl{W} \label{eq:dzda} \\
    \deriv{\al[m]}{\boldl{s}[n]} &= \deriv{ (\bm{\epsilon} * \boldl{s})[m]}{\boldl{s}[n]} = \deriv{\sum_{k=1}^{m} \epsilon_{m-k} \cdot \boldl{s}[k]}{\boldl{s}[n]} = \epsilon_{m-n} \cdot \bm{I}, \label{eq:dads}
\end{align}
where $\bm{I}$ denotes the identity matrix.
We then obtain
\begin{align}
    \el[n] :&= \deriv{\loss}{\al[n]}= \deriv{\loss}{\zlL[n]} \deriv{\zlL[n]}{\al[n]} = \boldx{d}{l+1}[n] \boldl{W} \label{eq:el}\\
    \boldl{d}[n] :&= \deriv{\loss}{\zl[n]}  = \sum_{m=n}^T \sum_{k=m}^T \deriv{\loss}{\al[k]} \deriv{\al[k]}{\boldl{s}[m]} \deriv{\boldl{s}[m]}{\zl[n]} \nonumber \\
    &= \sum_{m=n}^T \Big( \sum_{k=m}^T \epsilon_{k-m} \el[l] \Big) \sigl_n[m] = \sum_{m=n}^T \big( \bm{\epsilon} \odot \el \big)[m] \sigl_n[m] , \label{eq:dl}
\end{align}
where we used \eqref{eq:dzda} and \eqref{eq:dads} and where \(\odot\) denotes element-wise correlation operation in time. 
These equations are solved  for $\el[.]$ and $\dl[.]$ iteratively starting from \(\mathbf{e}^{(L)}[.] = \deriv{\loss}{\mathbf{a}^{(L)}[.]}\) at the output layer.

\iffalse
The gradients of the loss with respect to \(\wl_i \in \mathbb{R}^{N_{l}}\), the \(i\)-th row of weight matrix \(\Wl\), are
\begin{align*}
    \deriv{\loss}{\wl_i} &= \sum_{n=1}^T \deriv{\loss}{z^{(l+1)}_i[n]} \deriv{z^{(l+1)}_i[n]}{\wl_i} = \sum_{n=1}^T d^{(l+1)}_i[n] \cdot (\al)^ \top [n],
\end{align*}
where $\top$ denotes the transpose operation and where we used  \eqref{eq:z} and that only the \(i\)-th component of \(\zl\) depends on \(\wl_i\).
Therefore
\begin{align}\label{eq:gradw}
    \deriv{\loss}{\Wl} &= \sum_{n=1}^T \boldx{d}{l+1}[n] \cdot \al[n]^ \top.
\end{align}
\fi

The gradients of the loss with respect to weight matrix \(\Wl\), are then given by
\begin{align}\label{eq:gradw}
    \deriv{\loss}{\Wl} &= \sum_{n=1}^T \deriv{\loss}{\boldx{z}{l+1}[n]} \deriv{\boldx{z}{l+1}[n]}{\Wl} = \sum_{n=1}^T \boldx{d}{l+1}[n] \cdot \al[n]^ \top,
\end{align}
where $\top$ denotes the transpose operation and where we used  \eqref{eq:z}.

\subsubsection{Simplification for LIF and IF neurons}
\label{ss:lif}

Our derivation of the gradients in the previous section applies to an arbitrary \ac{srm}.
%spike response model (SRM).
Since many \acp{snn} are based on the \ac{lif} neuron model, in this section, we derive a more compact expression for this special case. 

Using the same notation as in Section~\ref{ss:srm}, the membrane potential in discrete-time is described by
\[u[n] = \alpha u[n-1] + \sum_i w_i s_i^{in}[n],\]
where \(\alpha := \exp{\frac{-\Delta}{\tau}} \in (0,1)\) is the decay factor in the LIF model determined by the membrane time constant \(\tau\) and simulation time step \(\Delta\).

The \ac{srm} of \ac{lif} dynamics is given by the spike response and reset kernels \(\epsilon_n = \alpha^n \mathbb{1}_{\{n\geq 0\}}\) and \(\nu_n = -\alpha^n \theta \mathbb{1}_{\{n\geq 0\}}\), where $\mathbb{1}$ denotes the indicator function and \(\theta\) the firing threshold.
The derivatives of \(\boldl{s}[.]\) in  \eqref{eq:dsdz} can be expressed in closed-form as follows
\begin{align}
    \sigl_m[n]  &= \begin{cases}\label{eq:dsdzlif}
        0 & n<m \\
        \boldl{f'}[n] & n=m \\
        - \theta \boldl{f'}[n] \boldl{f'}[m] \boldl{\chi}_m[n] & n>m
    \end{cases}\\
    \boldl{\chi}_m[n] :&= \begin{cases}
        \bm{I} & n=m+1 \\
        \prod_{k=m+1}^{n-1} (\alpha \bm{I} - \theta \boldl{f'}[k]) & n>m+1
    \end{cases}
    \label{eq:chi}
\end{align}
where $\bm{I}$ denotes the identity matrix.
We refer to Appendix~\ref{appdx:lif} for further details.
This yields a computationally efficient implementation in which   \(\boldl{\chi}_m\) are calculated iteratively as
\(\boldl{\chi}_m[n+1] = \boldl{\chi}_m[n] (\alpha \bm{I} - \theta \boldl{f'}[n])\).

This analysis can be applied equivalently to \ac{if} neurons without leak by setting the decay factor \(\alpha \equiv 1\).

\subsubsection{Comparison with SLAYER gradients}
\label{ss:slayer}
By comparing the gradients computed in Section~\ref{ss:grads} with those in SLAYER~\citep{shrestha_orchard18}, we find that the expressions for \(\el[.]\) and \(\deriv{\loss}{\Wl}\) (see equations \eqref{eq:el} and \eqref{eq:gradw}) match. 

The difference lies mainly  in \(\boldl{d}\) (equation \eqref{eq:dl}) and more specifically in the derivatives \(\deriv{\boldl{s}[n]}{\zl[m]}\), which are set to \(0\) for \(m \neq n\) in SLAYER. In particular, we would obtain the gradients in SLAYER by setting the reset kernel  to \(0\), where in that case \(\sigl_m[n]\) is only nonzero for \(m = n\) and \(\boldl{d}[n]\) becomes \(\boldl{f'}[n] \cdot \big( \bm{\epsilon} * \el \big)[n]\).

In the concrete case of LIF neuron dynamics (cf. Section~\ref{ss:lif}), we may indeed notice that the quality of the approximation of the gradients in SLAYER depends on how the matrix \(\boldl{\chi}_m\) (see Equation \eqref{eq:chi}) decays as a function of index-difference $|n-m-1|$. This decay, in general, can be characterized in terms of singular values of \(\boldl{f'}[n]\). 
However, since the matrix of surrogate gradients \(\boldl{f'}[n]\) are all diagonal, this boils down to how the diagonal elements 
\begin{align*}
\Big ( \boldl{\chi}_m [n] \Big )_{i,i}&= \prod_{k=m+1}^{n}(\alpha  - \theta f'_s(u_i^{(l)}[k]) )
\end{align*}
decay as a function of $n-m-1$ (for $n\geq m+1$). We may study several interesting scenarios:

(i)\,In the usual case where $f_s$ is an increasing function, $f_s'(u)$ is positive for all $u\in[0, \theta]$. If in addition, one designs the surrogate gradient $f'_s$ such that $f'_s(u) \in [0, \frac{\alpha-\mu}{\theta}]$ for some $\mu\in[0,\alpha]$ for all $u \in [0, \theta]$, one may obtain the bounds \(0\leq \Big ( \boldl{\chi}_m [n] \Big )_{i,i} \leq \mu^{n-m-1}\), which implies  that $\Big ( \boldl{\chi}_m [n] \Big )_{i,i}$ is quite small if $n\geq m+1 + O(\log \frac{1}{\mu} )$. As a result, compared with SLAYER, which assumes $\boldl{\chi}_m [n] \equiv 0$ for all $m, n, l$, our method takes into account additional $O(\log \frac{1}{\mu} )$ correction terms. 

(ii)\,If the surrogate gradient is not designed properly such that it is larger than  $\frac{\alpha}{\theta}$ for some range of $u$ within $[0, \theta]$, the term 
$\boldl{\chi}_m [n]$ may become significantly large as $n \gg m+1$. In such a case, we may expect a large deviation between the exact gradients derived from our method and the approximate gradients in SLAYER. We speculate that this may be the reason SLAYER gradients are more sensitive to the scaling of the surrogate gradients (cf. simulation results in Section~\ref{sec:simulation}).

\section{Simulation results}
\label{sec:simulation}

%\subsection{Benchmark Tasks}
We show that EXODUS leads to faster convergence when applied to different tasks, as our method computes the same gradients as BPTT in contrast to SLAYER. We benchmark EXODUS on 3 neuromorphic tasks with increasingly temporal features: (i)\,DVS Gesture~\citep{Amir_2017_CVPR}, (ii)\,Heidelberg Spiking Digits~\citep{cramer2020heidelberg}, and (iii)\,Spiking Speech Commands~\citep{cramer2020heidelberg}. To compare numerical stability, experiments are repeated over different gradient scaling factors \(f_s\) (see Eq.~\ref{eq:s}). When comparing EXODUS and SLAYER, we make sure that forward dynamics, random seed and initial weights are the same. We used SLAYER's official implementation available on Github. 
Detailed architecture and training parameters are described in the appendix in Table~\ref{tab:training_parameters}. 

\begin{figure}
    \centering
    \textbf{DVS Gesture} \\
    \includegraphics[width=0.29\textwidth]{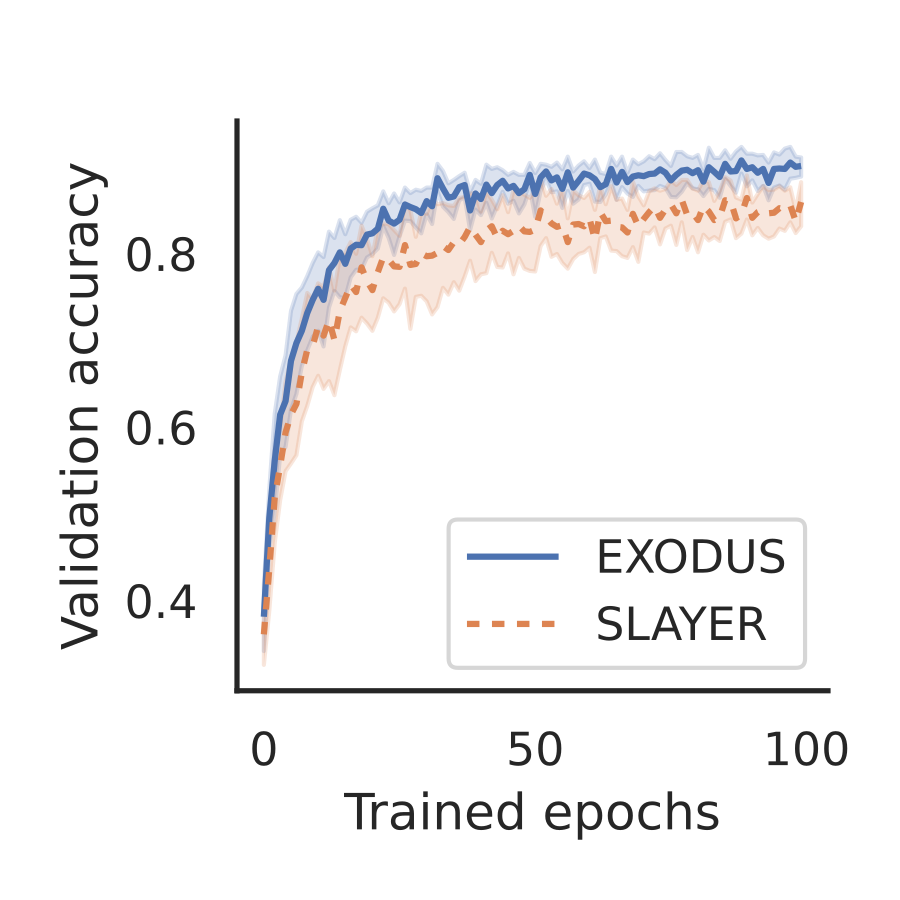}
    \includegraphics[width=0.45\textwidth]{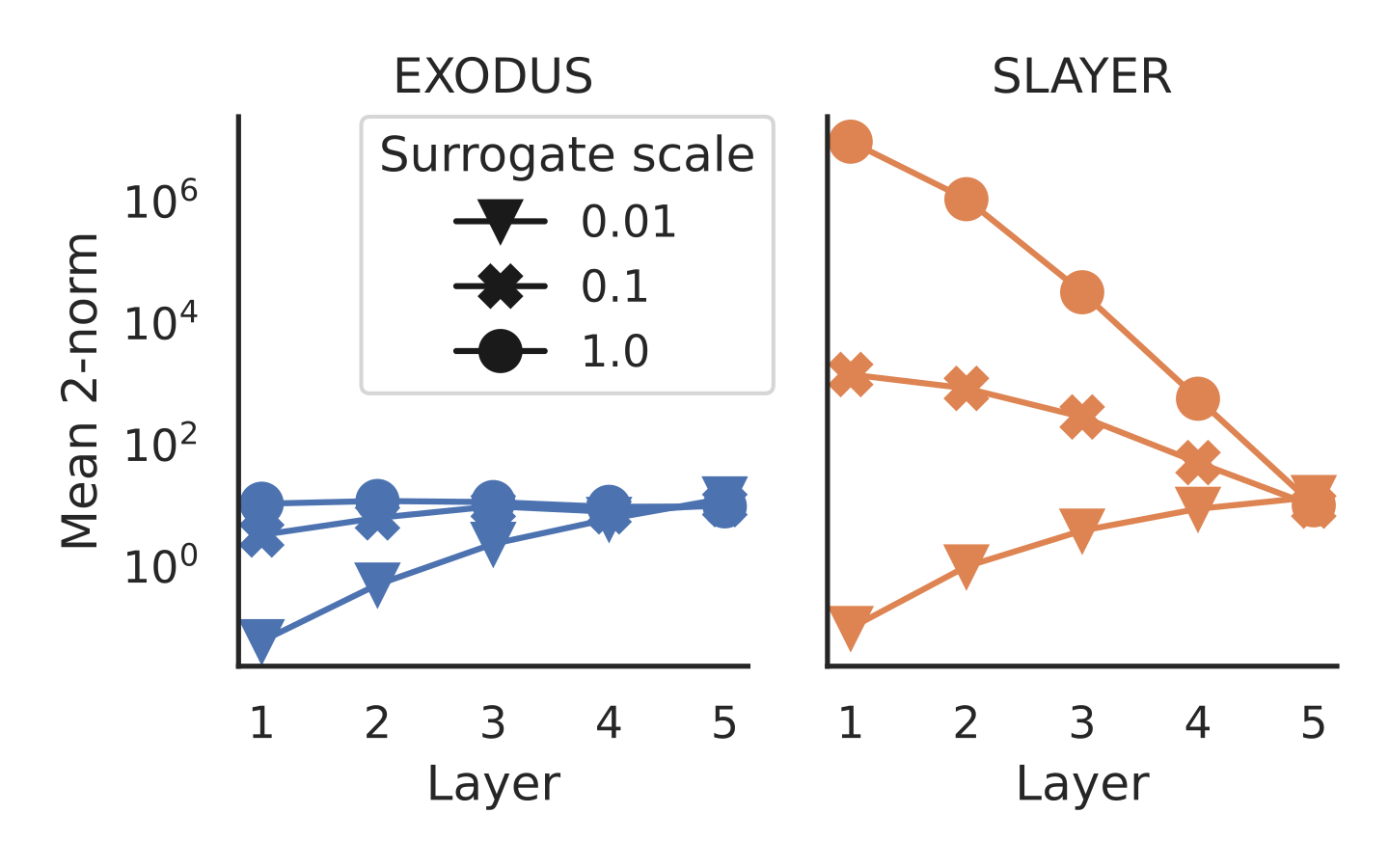}
    \includegraphics[width=0.22\textwidth]{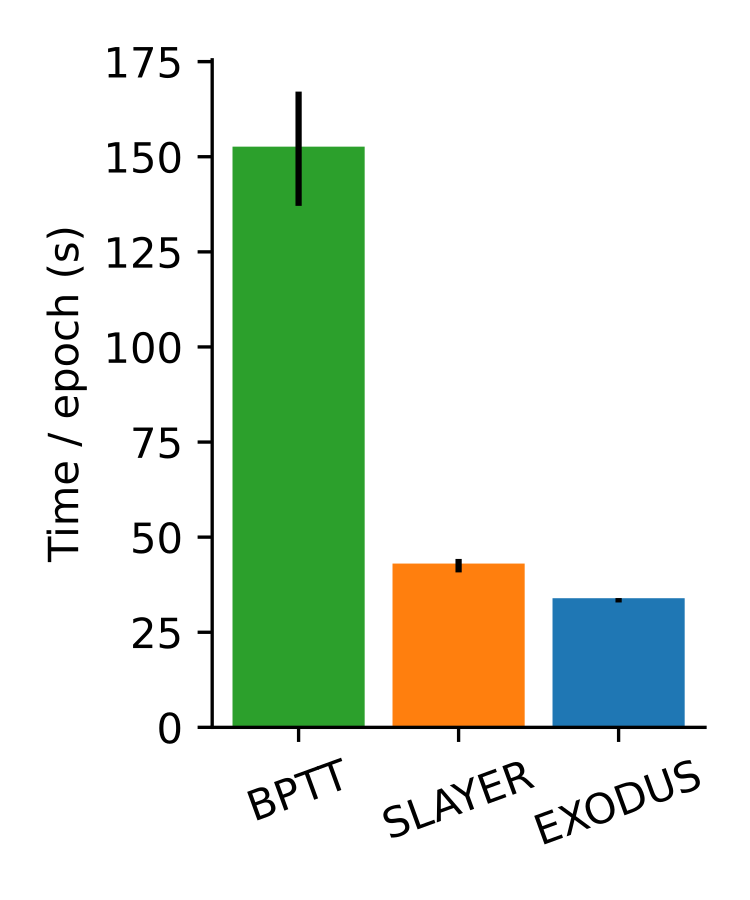}
    \textbf{Heidelberg~Spiking~Digits} \\
    \includegraphics[width=0.29\textwidth]{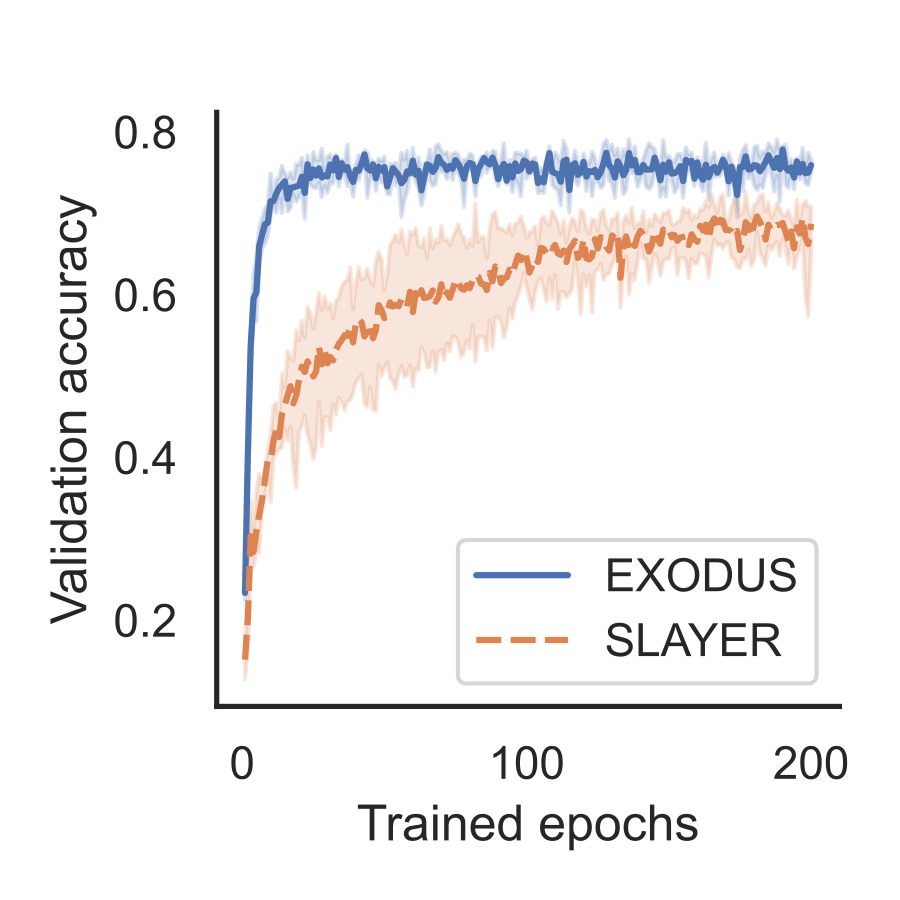}
    \includegraphics[width=0.45\textwidth]{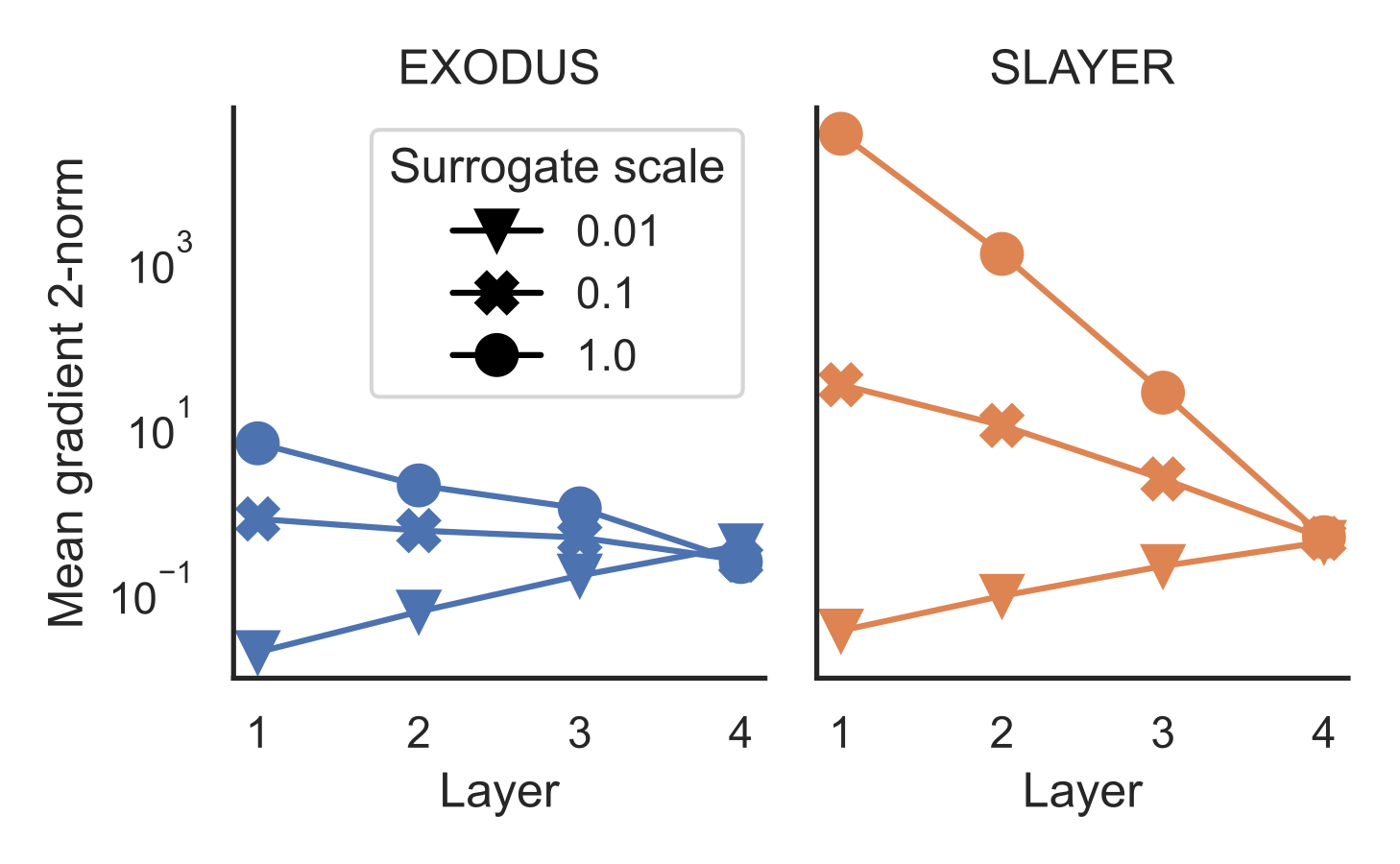}
    \includegraphics[width=0.22\textwidth]{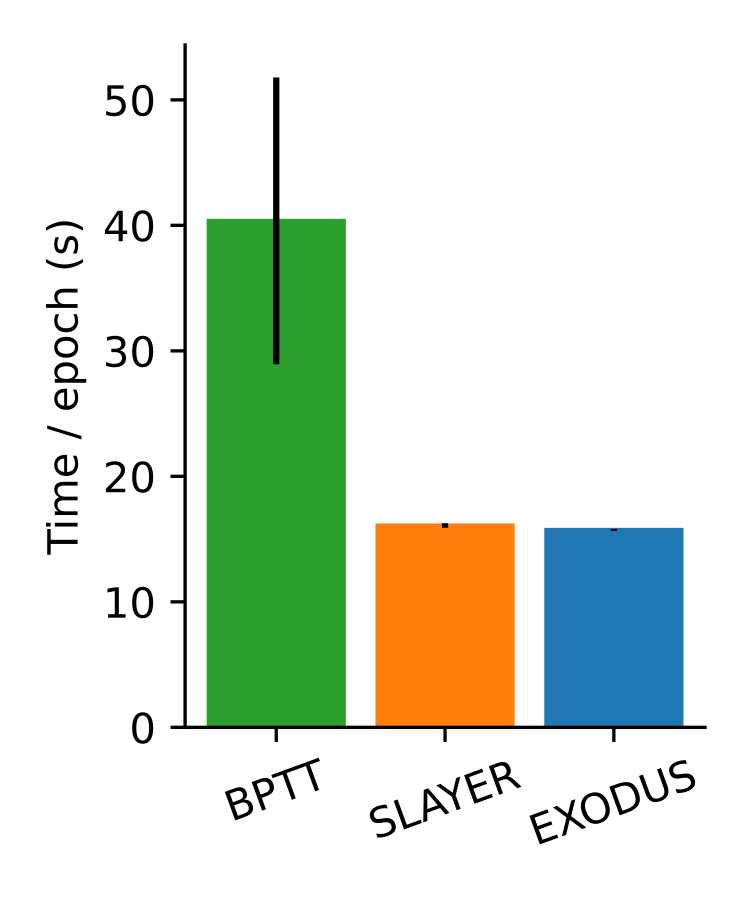}
    \textbf{Spiking~Speech~Commands} \\
    \includegraphics[width=0.29\textwidth]{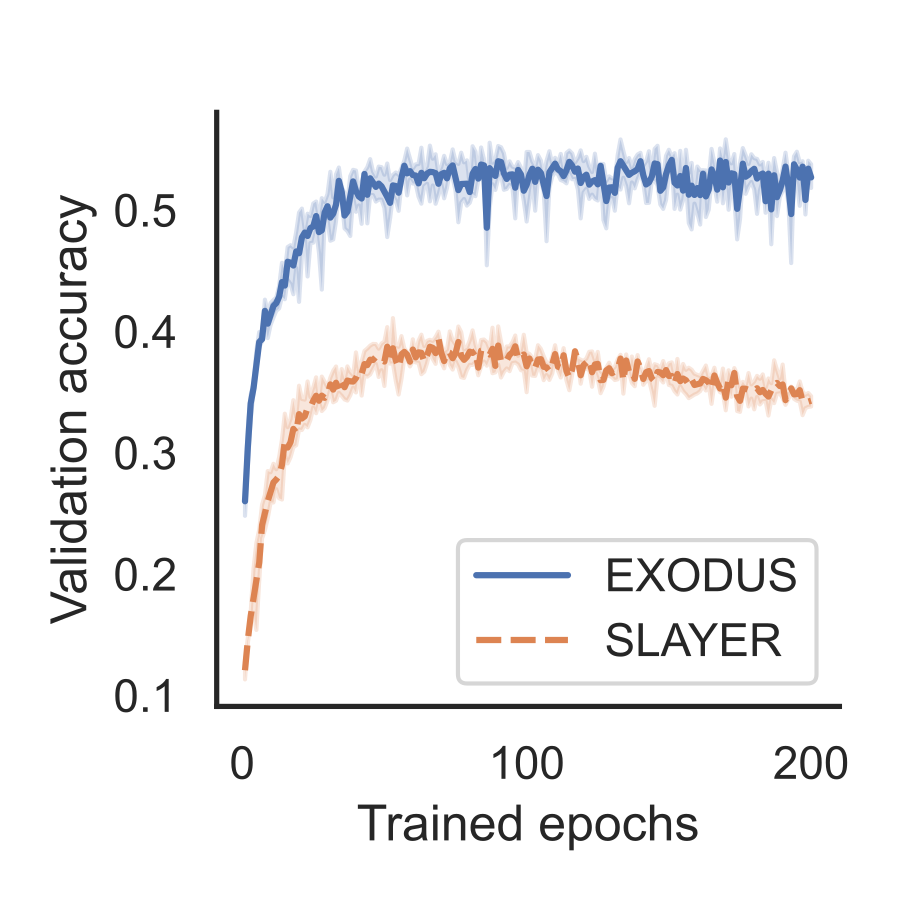}
    \includegraphics[width=0.45\textwidth]{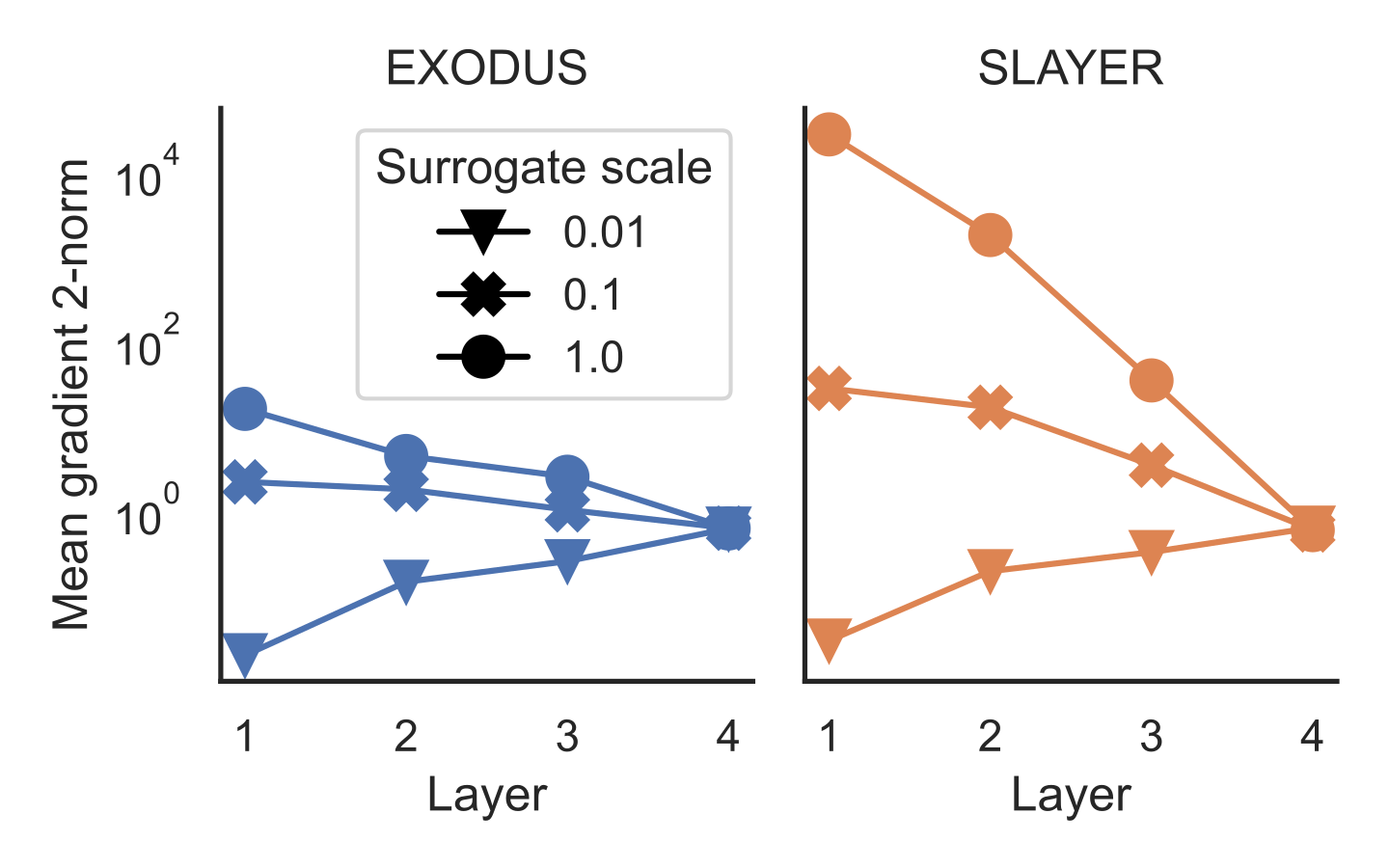}
    \includegraphics[width=0.22\textwidth]{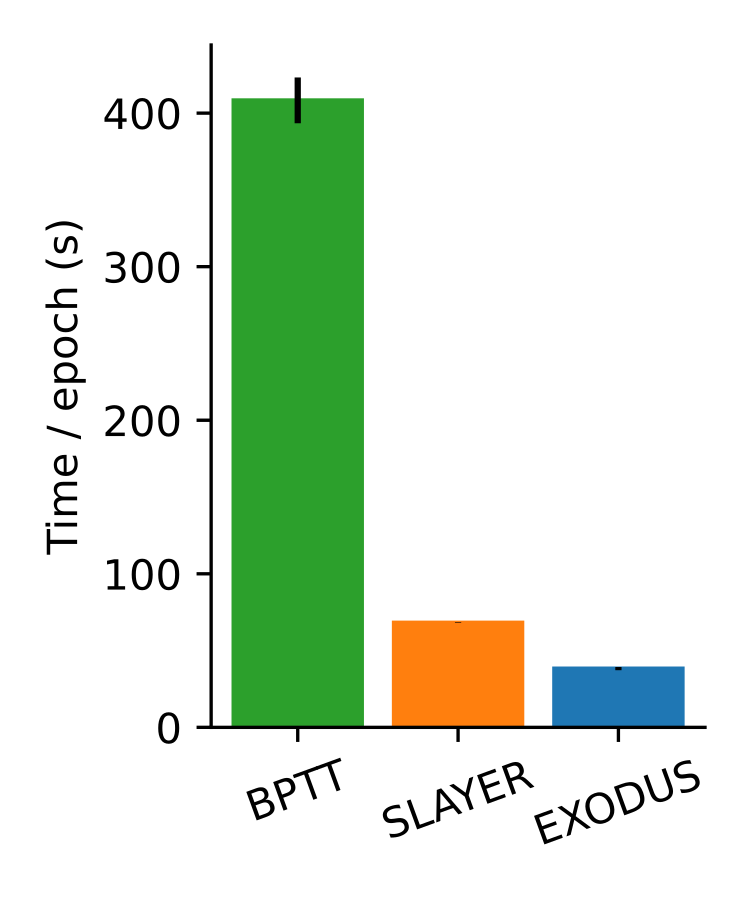}
    \caption{Performance on DVS Gesture (top row), Heidelberg Spiking Digits (center row) and Spiking Speech Commands (bottom row) datasets. \textbf{Left column:} Validation accuracy averaged over 3 runs for each algorithm using a gradient scale of $1$. 
    EXODUS (solid blue curves) generally converges much faster and reaches a higher accuracy than SLAYER (dashed orange curves), in particular for tasks with strong temporal dependencies.  
    \textbf{Center column:} Mean 2-norms of weight gradients during training, for individual layers and different scaling of surrogate gradient. When surrogate gradients are not down-scaled, gradients for SLAYER explode toward the input layer, whereas for EXODUS they remain mostly stable. Only for very low scaling, gradients vanish for both algorithms. 
    \textbf{Right column:} Training speed measured in wall clock time per epoch for different implementations. EXODUS is consistently the fastest.
    }
    \label{fig:dvs_performance}
\end{figure}

%\subsection{DVS Gesture}
%\label{ss:dvs_gesture}
\paragraph{DVS Gesture} is a dataset released under Creative Commons Attribution 4.0 license and consists of 1342 DVS camera~\citep{Lichtsteiner08} recordings of 29 subjects performing 11 hand and arm gestures under different illumination conditions~\citep{Amir_2017_CVPR}. Recordings from 23 subjects are designated as training set, the remaining 6 subjects for the test set. The task is to classify the gestures.
Similar to~\citet{shrestha_orchard18}, we  use only the first 1.5\,s of each video sequence, with a simulation time step of 5\,ms. To classify the gestures, we apply a spiking CNN architecture of four convolution layers and one fully connected layer, with Integrate-and-Fire neurons and cross entropy loss. 
%Camera events are streamed directly into the network, without any pre-processing. Training is performed with ADAM~\citep{kingma2014adam} as optimizer. 
The left-most columns in Fig.~\ref{fig:dvs_performance} and Tab.~\ref{tab:performance_numbers} show validation accuracies over the course of 100 training epochs. EXODUS reaches a higher accuracy than SLAYER (\((92.8 \pm 2.2)\%\) vs \((87.8 \pm 3.0) \%\)). The centre column in Fig.~\ref{fig:dvs_performance} shows how weight gradients for individual layers scale under different surrogate gradient scales. SLAYER is much more sensitive to the scale, with gradients often taking increasingly high values for layers that are further away from the output.

%\subsection{Heidelberg Spiking Digits}
%\label{ss:hsd}
\paragraph{Heidelberg Spiking Digits (HSD)} is an event-based audio classification dataset with highly temporal features, with samples recorded from a silicon cochlear~\citep{cramer2020heidelberg}. 
We used a 4-layer fully-connected architecture with Integrate-and-Fire neurons and max-over-time loss. As shown in Fig.~\ref{fig:dvs_performance} and Tab.~\ref{tab:performance_numbers}, training converges faster and reaches a significantly higher accuracy using EXODUS compared to SLAYER (\(78.01 \% \) vs \( 70.58 \% \)). Gradients are much more stable for earlier layers in EXODUS and a speedup of \(2.57\) is achieved when compared to BPTT.

%\subsection{Spiking Speech Commands}
\paragraph{Spiking speech commands (SSC)} is a neuromorphic version of Google's Speech Command dataset~\citep{cramer2020heidelberg}. 
We use the same 4-layer fully-connected architecture with Integrate-and-Fire neurons as for the HSD task. 
Here once again our simulation measurements in Fig.~\ref{fig:dvs_performance} and Tab.~\ref{tab:performance_numbers} show that gradients across layers in EXODUS are much more stable than when using SLAYER. Validation accuracy is much higher in this task using EXODUS ( \( 55.4 \% \) vs \( 40.1 \% \) for SLAYER) and also provides considerable speedup of \(10.67\) when compared to our BPTT implementation.

\paragraph{Training Speed} is shown in the right-most column in Fig.~\ref{fig:dvs_performance} and in Tab.~\ref{tab:performance_numbers}. It shows training speeds measured for the three datasets and three implementations. All speed tests are executed on a NVIDIA GeForce 1080 Ti. Our BPTT implementation is based on PyTorch 1.11. Across the three datasets, SLAYER is between \(2.51\) and \(5.99\) times faster than our BPTT implementation, whereas EXODUS is \(2.57\) to \(10.67\) times faster. Raw duration numbers measured are provided in the appendix, Tab.~\ref{tab:training_time_seconds}. It should be noted that the duration includes data loading, forward and backward pass as well as the optimizer update.

\begin{table}
    \centering
    \begin{tabular}{cccccc}
        \toprule
            & \multicolumn{2}{c}{Validation accuracy [\%]} && \multicolumn{2}{c}{\makecell{Mean training time\\speedup compared to BPTT}} \\
            & EXODUS & SLAYER && EXODUS & SLAYER \\
        \midrule
        DVS & $\mathbf{92.8\pm2.2}$   & $87.8\pm3.0$   && $\mathbf{4.55\times}$  & $3.58\times$ \\
        HSD & $\mathbf{78.01\pm0.2}$ & $70.58\pm1.9$ && $\mathbf{2.57\times}$  & $2.51\times$ \\
        SSC & $\mathbf{55.41\pm0.4}$ & $40.1\pm0.8$  && $\mathbf{10.67\times}$ & $5.99\times$ \\
        \bottomrule
    \end{tabular}
    \caption{Validation accuracy and training time per epoch for different datasets. Validation accuracy is mean and standard deviation of maximum accuracies across 3 runs with a gradient scale of $1$. Training time is measured per epoch on a NVIDIA GeForce 1080 Ti averaged across 3 epochs. Raw numbers for training time are given in the appendix, Tab.~\ref{tab:training_time_seconds}.}
    \label{tab:performance_numbers}
\end{table}

\section{Discussion}
\label{sec:discussion}
We have shown that while SLAYER enables training of spiking neural networks at high computational efficiency, it does so at the cost of omitting the effect of the neuron's reset mechanism on the gradients. With our newly proposed algorithm EXODUS, we present a modification of SLAYER that computes the same gradients as those in the original BPTT while maintaining the high computational efficiency of SLAYER.
Tuning the surrogate gradient function is important when training deeper architectures~\citep{ledinauskas2020training}, which can be a costly manual process when using SLAYER. EXODUS makes training deep SNN architectures from scratch easier by providing less sensitivity to the gradient scale hyper parameter. Not only does the gradient magnitude scale in a stable manner but also the gradient direction leads to faster convergence as shown in our experiments. 
The difference is especially noticeable for long time constants and Integrate-and-Fire neurons (without any leak) in the extreme case, but also persists for shorter time constants (see Appendix~\ref{sec:poisson_fitting}). We argue that because of the neuron's longer memory the contribution of the neuron's reset mechanism increases, which is not taken into account in SLAYER's gradient computation. 
For Integrate-and-Fire neurons, both with or without leak, the terms that ensure correct representation of the reset mechanism in the gradients with EXODUS allow for a computationally efficient implementation, resulting in similar computation speeds as SLAYER. It should be noted, however, that for other neuron dynamics this may not always be possible.
Furthermore, our work draws attention to the fact that arriving at mathematically correct gradients is not always trivial and that inaccuracies are sometimes hard to spot in the literature. Our method of deriving gradients through the \ac{ift}, when no explicit functional relation exists between two or more variables, is of independent interest for devising new learning strategies in a rigorous manner.

% We are convinced that this paper is an important contribution to the field of spiking neural networks by providing an algorithm for efficient and stable learning with BPTT.

%%%%%%%%%%%%%%%%%%%%%%%%%%%%%%%%%%%%%%%%%%%%%%%%%%%%%%%%%%%%

\bibliographystyle{plainnat}
\bibliography{references}

%%%%%%%%%%%%%%%%%%%%%%%%%%%%%%%%%%%%%%%%%%%%%%%%%%%%%%%%%%%%
\newpage

\section{Appendix}

\subsection{Examples for application of Implicit Function Theorem}
\label{appdx:ift_examples}

\begin{figure}[ht]
\centering
\includegraphics[width=0.5\columnwidth]{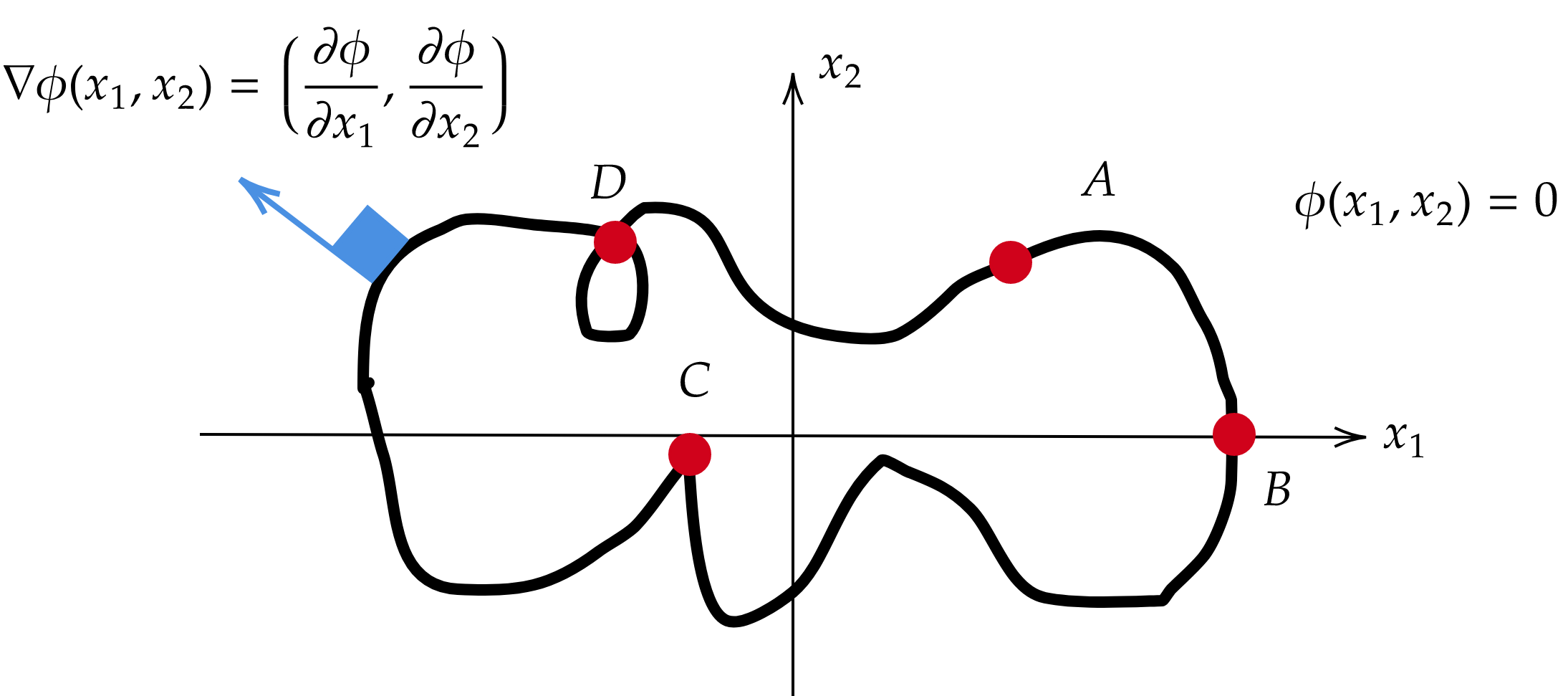}
\caption{Illustration of the implicit function theorem.}
\label{fig_ift}
\end{figure}

{\bf Example 1.} Fig.\,\ref{fig_ift} illustrates the zero-set $\{(x_1,x_2): \phi(x_1,x_2)=0\}$ of a function $\phi: \mathbb{R}^2 \to \mathbb{R}$. To investigate the conditions of IFT, we first note that the gradient of $\phi$ denoted by $\nabla \phi=(\frac{\partial \phi}{\partial x_1}, \frac{\partial \phi}{\partial x_2})$  is always orthogonal to the level-set (here the zero-set) of $\phi$. Thus, by observing the orthogonal vector to the level-set, we can verify if $\frac{\partial \phi}{\partial x_1}$ or $\frac{\partial \phi}{\partial x_2}$ are non-singular (non-zero in the scalar case we consider here).
We consider several case:
\begin{itemize}
    \item point $D$ and $C$: at these two points the gradient vector does not exist, so the assumptions of the IFT are not fulfilled. However, we can see that at point $C$, one can still write $x_2$ as function of $x_1$ although this function is not differentiable.
    
    \item point $A$: gradient vector has nonzero vertical and horizontal components, i.e., $\frac{\partial \phi}{\partial x_1} \not = 0$ and $\frac{\partial \phi}{\partial x_2} \not = 0$. Thus, from IFT, one should be able to write both $x_1$ and $x_2$ locally as a differentiable function of the other.
    
    \item point $B$: here $\frac{\partial \phi}{\partial x_1} \not = 0$ and $x_1$ is locally a differentiable function of $x_2$. However, since $\frac{\partial \phi}{\partial x_2} = 0$, the assumptions of IFT do not hold for $x_2$. And, it is seen from Fig.\,\ref{fig_ift} that $x_2$ cannot be written as function of $x_1$ in an open neighborhood of $B$.
\end{itemize}

One of the implications of the IFT is that, given an implicit functional relationship between a collection of variables, we are not allowed to apply partial differentiation and chain rule in an ad-hoc fashion. Otherwise, we may obtain wrong results. We illustrate this by the following example.

{\bf Example 2. (chain rule over a loopy graph)} Let $\phi: \mathbb{R}^3 \to \mathbb{R}$ be a differentiable function and let $(x_1,x_2,x_3)$ be a point in the zero-set $\phi(x_1,x_2,x_3)=0$.
By an ad-hoc application of the chain rule, one may start from $x_1=x_1$ and apply the chain rule to obtain 
\begin{align}
    1=\frac{\partial x_1}{\partial x_1}= \frac{\partial x_1}{\partial x_2} \times \frac{\partial x_2}{\partial x_3} \times \frac{\partial x_3}{\partial x_1}.\label{ift_3var_ex_1}
\end{align}
Now let us go step-by-step using the IFT and show that this result is indeed wrong.
Let us consider computation of $\frac{\partial x_1}{\partial x_2}$. This computation simply implies that $x_1$ and $x_2$ should be treated as dependent and independent variables, respectively. Since we have only a single equation $\phi(x_1,x_2,x_3)=0$, we can solve one of the variables as a function of the remain two variables. So, overall, we need to treat $x_1$ and $(x_2,x_3)$ as dependent and independent variables, respectively.
Now we can apply the IFT to obtain
\begin{align}
    \frac{\partial \phi}{\partial x_1}\times \frac{\partial x_1}{\partial x_2} + \frac{\partial \phi}{\partial x_2}=0,
\end{align}
which implies that
\begin{align}
    \frac{\partial x_1}{\partial x_2}=-\frac{ \phi_2 } { \phi_1  },
\end{align}
where we defined the short-hand notation $\phi_i= \frac{\partial \phi}{\partial x_i}$, where we assumed that all the partial derivatives are evaluated at the point $(x_1,x_2,x_3)$, and where we assumed that all the partial derivative $\phi_i$ are non-zero.

Applying the symmetry, we therefore obtain that
\begin{align}
    \frac{\partial x_1}{\partial x_2} \times \frac{\partial x_2}{\partial x_3} \times \frac{\partial x_3}{\partial x_1} = -\frac{ \phi_2 } { \phi_1  } \times -\frac{ \phi_3 } { \phi_2  } \times -\frac{ \phi_1 } { \phi_3  } =-1.
\end{align}
Comparing this with \eqref{ift_3var_ex_1} simply shows that  ad-hoc application of the chain rule, especially when there is a \emph{loopy structure} as in $x_1\to x_2 \to x_3 \to x_1$,  yields a wrong result. 

\subsection{Detailed derivation of gradients}
\label{appdx:sol}

In the following, we provide a detailed step-by-step derivation of the the derivatives \(\deriv{\boldl{s}[n]}{\boldl{z}[m]}\) as defined in the main document in Section \ref{ss:grads}. Since the derivation is similar for different layers, for simplicity, we drop the superscript \(^{(l)}\) here.

Let us vectorize the set of equations at a specific layer as $\bm{\Phi}=[\bm{\Phi}_s, \bm{\Phi}_u]^\top$ where $\bm{\Phi}_u=\{\bm{\Phi}_u[n]: n\in [T]\}$ and $\bm{\Phi}_s=\{\bm{\Phi}_s[n]: n\in [T]\}$. Similarly let us vectorize the dependent variables as $\mathcal{D}=[\mathcal{D}_u, \mathcal{D}_s]^\top$ where $\mathcal{D}_u=\{\mathbf{u}[n]: n \in [T]\}$ and $\mathcal{D}_s=\{\mathbf{s}[n]: n \in [T]\}$.
    
To verify the conditions of \ac{ift}, we can check that all the equations are differentiable functions of all the variables, provided that $f_s$ is a differentiable function. In addition, we need to verify that the Jacobian matrix $\mathbf{J}^{\mathcal{D}}$ of the partial derivatives of \(\bm{\Phi}\) with respect to \(\mathcal{D}\) is non-singular. 

We can write it as:
\begin{align*}
    \mathbf{J}^{\mathcal{D}} &= \deriv{\bm{\Phi}}{\mathcal{D}} =
    \begin{pmatrix}
        \deriv{\bm{\Phi_s}}{\bm{s}} & \deriv{\bm{\Phi_s}}{\bm{u}} \\
        \deriv{\bm{\Phi_u}}{\bm{s}} & \deriv{\bm{\Phi_u}}{\bm{u}} \\
    \end{pmatrix}
    = \begin{pmatrix}
        \bm{I}_{N_l T} & -\mathbf{f'} \\
        \bm{\nu}_{-1} & \bm{I}_{N_l T} \\
    \end{pmatrix},
\end{align*}
where \(\bm{I}_{N_lT}\) denotes the identity matrix of order $N_l T$, and where we define
\begin{align*}
    \mathbf{f'} := 
    \begin{pmatrix}
            -\bm{f'}[1] & 0           & \dots  & 0  \\
            0           & -\bm{f'}[2] & \dots  & 0  \\
            \vdots      &             & \ddots & \vdots \\
            0           & 0           & \dots  & -\bm{f'}[T] \\
    \end{pmatrix},
\end{align*}
with \(\bm{f'}[n]\) denoting the diagonal matrix of the surrogate gradients \(\{f_s'(u_i[n]): i\in[N_l]\}\) at time instant \(n\), and where we define
\begin{align*}
    \bm{\nu}_{-1} := 
    \begin{pmatrix}
            0                           & 0                           & \dots  & 0      \\
            -\nu_0 \bm{I}_{N_l}     & 0                           & \dots  & 0      \\
            \vdots                      &                             & \ddots & \vdots \\
            -\nu_{T-2} \bm{I}_{N_l} & -\nu_{T-3} \bm{I}_{N_l} & \dots  & 0      \\    
    \end{pmatrix}.
\end{align*}

\iffalse
    &= \begin{pmatrix}
    \bm{I} & 0          & \dots  & 0          & | & -\bm{f'}[1] & 0              & \dots  & 0  \\
    0          & \bm{I} & \dots  & 0          & | & 0              & -\bm{f'}[2] & \dots  & 0  \\
    \vdots     &            & \ddots & \vdots     & | & \vdots         &                & \ddots & \vdots \\
    0          & 0          & \dots  & \bm{I} & | & 0              & 0              & \dots  & -\bm{f'}[T] \\
    \hline
    0                     & 0                     & \dots  & 0      & | & \bm{I} & 0          & \dots  & 0 \\
    -\nu_0 \bm{I}     & 0                     & \dots  & 0      & | & 0          & \bm{I} & \dots  & 0  \\
    \vdots                &                       & \ddots & \vdots & | & \vdots     &            & \ddots & \vdots \\
    -\nu_{T-2} \bm{I} & -\nu_{T-3} \bm{I} & \dots  & 0      & | & 0          & 0          & \dots  & \bm{I} \\
    \end{pmatrix}
\fi

Since applying simple column-wise operations will not change the invertibility of $\mathbf{J}^{\mathcal{D}}$, 
so we add a  $\bm{\nu}_{-1}$ multiple of the second column-block to the first column-block and obtain 
\begin{align*}
    \widetilde{\mathbf{J}}^{\mathcal{D}} = 
        \begin{pmatrix}
            \bm{I}_{N_l T} -  \bm{\nu}_{-1} \mathbf{f'} & 0  \\
            \bm{\nu}_{-1}                                  & \bm{I}_{N_l T} \\
        \end{pmatrix}.
\end{align*} 
Because \(\bm{\nu}_{-1}\) is a lower triangular matrix with \(0\)-diagonal and $\mathbf{f'}$ is a diagonal matrix, $\bm{\nu}_{-1} \mathbf{f'}$ will be a lower triangular matrix with \(0\)-diagonal. This implies that \(\widetilde{\mathbf{J}}^{\mathcal{D}}\) is also lower triangular with all diagonal entries equal to \(1\). The determinant of such a matrix is \(1\), which proves the invertibility of \(\mathbf{J}^{\mathcal{D}}\) and verifies the non-singularity condition needed in \ac{ift}.

Since the conditions of \ac{ift} are fulfilled, we know that all the dependent variables \(\mathcal{D}\) are differentiable w.r.t. independent variables \(\mathcal{I}=\{\mathbf{z}[n]: n \in [T]\}\).

We find the derivatives \(\mathbf{D} = \deriv{\mathcal{D}}{\mathcal{I}} \) by solving 
\[\mathbf{J}^{\mathcal{D}} \cdot \mathbf{D} = - \mathbf{J}^{\mathcal{I}},\]
where the Jacobian w.r.t. the independent variables is
\begin{align*}
    \mathbf{J}^{\mathcal{I}} = \deriv{\bm{\Phi}}{\mathcal{I}} = 
    \begin{pmatrix}
        \deriv{\bm{\Phi_s}}{\bm{z}}\\
        \deriv{\bm{\Phi_u}}{\bm{z}}\\
    \end{pmatrix} =
    \begin{pmatrix}
        0 \\
        - \bm{I}_{N_l T} \\
    \end{pmatrix}.
\end{align*}

Applying the same row-wise operations to \(\mathbf{J}^{\mathcal{I}}\) as to \(\mathbf{J}^{\mathcal{D}}\), we get
\begin{align*}
    \widetilde{\mathbf{J}}^{\mathcal{I}} = 
        \begin{pmatrix}
            - \mathbf{f'} \\
            - \bm{I}_{N_l T} \\
        \end{pmatrix}.
\end{align*} 
To find \(\mathbf{D}\), we can now solve:
\begin{align*}
    \widetilde{\mathbf{J}}^{\mathcal{D}} \cdot \mathbf{D} &= - \widetilde{\mathbf{J}}^{\mathcal{I}} \\
    \Leftrightarrow \begin{pmatrix}
        \bm{I}_{N_l T} - \mathbf{f'} \bm{\nu}_{-1} & 0  \\
        \bm{\nu}_{-1}                                  & \bm{I}_{N_l T} \\
    \end{pmatrix} \cdot
    \begin{pmatrix}
        \deriv{\bm{s}}{\bm{z}}\\
        \deriv{\bm{u}}{\bm{z}}\\
    \end{pmatrix} &=
    \begin{pmatrix}
        \mathbf{f'} \\
        \bm{I}_{N_l T} \\
    \end{pmatrix}.
\end{align*}
Specifically, for \(\deriv{\bm{s}}{\bm{z}}\), we find,
\begin{align}
    \big( \bm{I}_{N_l T} - \mathbf{f'} \cdot \bm{\nu}_{-1} \big) \cdot \deriv{\bm{s}}{\bm{z}} = \mathbf{f'}. \label{eq:dsdzmatr}
\end{align}
Because \(\bm{I}_{N_l T} - \mathbf{f'} \cdot \bm{\nu}_{-1}\) has lower triangular shape, we can solve equation \eqref{eq:dsdzmatr} through forward substitution, yielding for any \(m, n \in [T]\) 
\begin{align}
    \deriv{\bm{s}[n]}{\bm{z}[m]} - \mathbf{f'}[n] \sum_{k=1}^{n-1} \nu_{n-1-k} \deriv{\bm{s}[k]}{\bm{z}[m]} &= \delta_{m,n} \mathbf{f'}[n] 
    \ \ \Rightarrow \ \  \deriv{\bm{s}[n]}{\bm{z}[m]} = \delta_{m,n} \mathbf{f'}[n] + \mathbf{f'}[n] \sum_{k=1}^{n-1} \nu_{n-1-k} \deriv{\bm{s}[k]}{\bm{z}[m]}. \label{eq:dsdzsol}
\end{align}

Starting with \(n=1\) and applying induction on $n$ and $m$, and using the right-hand-side expression for $\deriv{\bm{s}[n]}{\bm{z}[m]}$ in \eqref{eq:dsdzsol}, it is not difficult to verify that 
\begin{align}
    \deriv{\bm{s}[n]}{\bm{z}[m]}=0, \text{  for  \( m > n\)}.\label{eq:dsdzsol_step1}
\end{align}
Using this result and replacing  \(n=m\) in \eqref{eq:dsdzsol} yields 
\begin{align}
    \deriv{\bm{s}[m]}{\bm{z}[m]} = \mathbf{f'}[m].\label{eq:dsdzsol_step2}
\end{align}
Finally, replacing \eqref{eq:dsdzsol_step1} and \eqref{eq:dsdzsol_step2} in \eqref{eq:dsdzsol}, we obtain for  \(m<n\)
\[ \deriv{\bm{s}[n]}{\bm{z}[m]} = \mathbf{f'}[n] \cdot \sum_{k=m}^{n-1} \nu_{n-1-k} \deriv{\bm{s}[k]}{\bm{z}[m]} =  \mathbf{f'}[n] \sum_{k=m}^{n-1} \nu_{n-1-k} \deriv{\bm{s}[k]}{\bm{z}[m]} . \]

Introducing the short-hand notation \(\bm{\sigma}_m[n] := \deriv{\bm{s}[n]}{\bm{z}[m]}\) and rewriting the sum as a convolution operation, we conclude:
\begin{align}
    \bm{\sigma}_m[n] = \begin{cases}
        0 & n<m \\
        \bm{f'}[n] & n=m \\
        \bm{f'}[n] \cdot \big(\bm{\nu} \ast \bm{\sigma}_m \big) [n-1] & n>m . \label{eq:dsdzshort}
    \end{cases}
\end{align}

\subsection{Gradients for Leaky Integrate-and-Fire model}\label{appdx:lif}

The \acf{lif} neuron model can be seen as a special case of the \acf{srm}, with the spike response and reset kernels \(\epsilon\) and \(\nu\) given as\footnote{Note that the spike response kernel \(\epsilon\) given here corresponds to the case where neural dynamics are modeled fully as exponential membrane decay. Synaptic dynamics could be included by replacing \(\epsilon\) with \(\tilde{\epsilon} := \epsilon * \rho\), with \(\rho\) being the synaptic impulse response. The validity of the following analysis would not be affected. }
\begin{align}
    \epsilon_n &= \alpha^n \mathbb{1}_{\{n\geq 0\}}\nonumber \\
    \nu_n &= -\alpha^n \theta \mathbb{1}_{\{n\geq 0\}}, \label{eq:nulif}
\end{align}
which allows us to find a slightly simpler formulation for \(\boldl{\sigma}_m[n] := \deriv{\boldl{s}[n]}{\boldl{z}[m]}\). Dropping the superscript \(^{(l)}\), the derivatives in their general form are
\begin{align*}
    \bm{\sigma}_m[n] = \begin{cases}
        0 & n<m \\
        \bm{f'}[n] & n=m \\
        \bm{f'}[n] \big( \bm{\nu} \ast \bm{\sigma}_m \big) [n-1] & n>m .
    \end{cases}
\end{align*}

Let us introduce a new variable
\begin{align*}
    \bm{\gamma}_m[n] := \begin{cases}
        0 & n<m \\
        \bm{I} & n=m \\
        \big(\bm{\nu} \ast \bm{\sigma}_m \big) [n-1] & n>m, % \label{eq:gma}
    \end{cases}
\end{align*}
such that \(\bm{\sigma}_m[n] = \bm{f'}[n] \bm{\gamma}_m[n] \). We first prove the following proposition.  

\begin{proposition}
With \(\nu\) as in equation \eqref{eq:nulif} and for \(m > 1, n > m+1\),
\begin{align}
    \bm{\gamma}_m[n] = - \theta \bm{f'}[m] \prod_{k=m+1}^{n-1} (\alpha \bm{I} - \theta \bm{f'}[k] ). \label{eq:proposition}
\end{align}
\end{proposition}

\begin{proof}
We prove this by applying induction on $n$ and $m$.
It is helpful to note that for \(n > m\), \(\bm{\gamma}_m[n]\) can be written as
\begin{equation}\label{eq:gamma_alpha}
    \bm{\gamma}_m[n] = \sum_{k=m}^{n-1} \nu_{n-1-k} \bm{\sigma}_m[k] = - \theta \sum_{k=m}^{n-1} \alpha^{n-1-k} \bm{\sigma}_m[k].
\end{equation}

Let us consider any \(m \geq 1\) and let us assume for now that there exists an \(n > m + 1\) for which the proposition holds. Then for \(n + 1\) we find from \eqref{eq:gamma_alpha} that
\begin{align*}
    \bm{\gamma}_m[n+1] &= - \theta \sum_{k=m}^{n} \alpha^{n-k} \bm{\sigma}_m[k] \\
    &= - \big( \theta \bm{\sigma}_m[n] + \theta \alpha \sum_{k=m}^{n-1} \alpha^{n-1-k} \bm{\sigma}_m[k]  \big) \\
    &= - \big( \theta \bm{\gamma}_m[n] \bm{f'}[n] - \alpha \bm{\gamma}_m[n] \big) \\
    &= - \bm{\gamma}_m[n] \big( \theta \bm{f'}[n] - \alpha \bm{I} \big) \\
    &\stackrel{(i)}{=} - \bm{f'}[m] \Big( \theta \prod_{k=m+1}^{n-1} \big( \alpha \bm{I} - \theta \bm{f'}[k] \big) \Big) \cdot \big( \alpha \bm{I}  - \theta \bm{f'}[n] \big) \\
    &= - \theta \bm{f'}[m] \prod_{k=m+1}^{n} (\alpha \bm{I} - \theta \bm{f'}[k] ),
\end{align*}
where in $(i)$, we applied the induction hypothesis for the given \(n\) and \(m\). 
Up to now, we have shown that if the result is true for a given $n$ and $m$ with $n>m+1$, it is true for all $n>m+1$. To complete the induction, therefore, we need to extend the result from $m$ to $m+1$. We verify this directly.

By definition of \(\bm{\gamma}\), for \(n = m + 1\), we have that \(\bm{\gamma}_m[m+1] = - \theta \alpha ^0 \bm{\sigma}_m[m] = - \theta \bm{f'}[m] \) and therefore \(\bm{\sigma}_m[m+1] = - \theta \bm{f'}[m] \bm{f'}[m+1]\). We can then show that for \(n = m + 2\) the proposition is true:
\begin{align*}
    \mathbf{\bm{\gamma}}_m[m+2] &= - \theta \alpha ^ 1 \bm{\sigma}_m[m] - \theta \alpha ^ 0 \bm{\sigma}_m[m+1] \\
    &= - \theta \big( \alpha \bm{f'}[m] - \theta \bm{f'}[m] \bm{f'}[m+1] \big) \\
    &= - \theta \bm{f'}[m] \big( \alpha \bm{I} - \theta \bm{f'}[m+1] \big).
\end{align*}

Hence the proposition holds for \(n = m+2\) and therefore for all \(n \geq m+2\), independent of the choice of \(m\).
\end{proof}

With \(\bm{\sigma}_m[n] = \bm{f'}[n] \bm{\gamma}_m[n] \) we can therefore write \(\bm{\sigma}\) as:
\begin{align*}
    \bm{\sigma}_m[n] = \begin{cases}
        0 & n<m \\
        \bm{f'}[m] & n=m \\
        -\theta \bm{f'}[m] \bm{f'}[m+1] & n = m+1 \\
        -\theta \bm{f'}[m] \bm{f'}[n] \prod_{k=m+1}^{n-1} \big( \alpha \bm{I} - \theta \bm{f'}[k] \big) & n>m+1 .
    \end{cases}
\end{align*}

\subsection{Training parameters}

We list the parameters that we trained our networks with. In architecture, first element is the input dimensions for one time step, a layer of 16c5 is a convolutional layer with 16 channels and a kernel size of 5 and a layer of 10l would be a simple linear layer with 10 output features. Between all weight layers we employ IF spiking layers to generate non-linear output.

\begin{table}[h]
    \caption{Training parameters for classification tasks. }
    \centering
    \begin{tabular}{llll}
        \toprule
                        & DVS Gesture & HSD & SSC \\
        \midrule
        sequence length & 300 & 250 & 250 \\
        architecture    & \makecell{64x64x2-2c3-2p-4c3-\\2p-8c3-2p-16c3-11l} & 100-128l-128l-20l & 100-128l-128l-35l \\
        optimiser       & ADAM & ADAM & ADAM \\
        loss            & sum over time CE & max over time CE & max over time CE \\
        learning rate   & 1e-3 & 1e-3 & 1e-3  \\
        mini-batch size & 32 & 128 & 128 \\
        epochs          & 100 & 200 & 200 \\
        \bottomrule
    \end{tabular}
    \label{tab:training_parameters}
\end{table}

\subsection{Poisson spike train fitting}
\label{sec:poisson_fitting}
\begin{figure}[!ht]
    \centering
    \includegraphics[width=0.5\columnwidth]{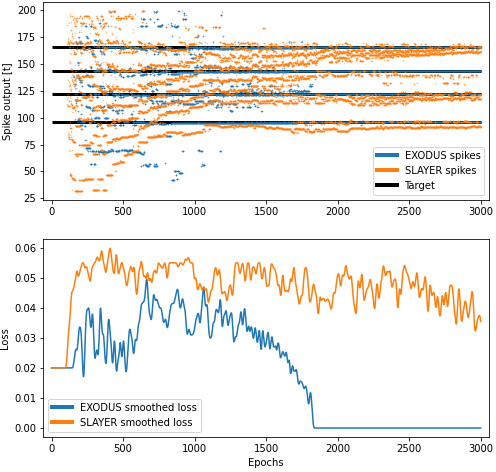}
    \includegraphics[width=0.45\columnwidth]{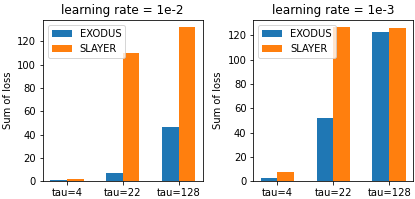}
    \caption{Poisson spike train fitting. \textbf{Upper:} spike output for networks when trained with EXODUS and SLAYER as well as target spike train, all tracked across epochs. \textbf{Lower:} Loss over time for example experiment. EXODUS shows faster convergence than SLAYER. \textbf{Right:} Sum of averaged loss across 5 experiments for one parameter combination (learning rate, LIF time constant tau) as well as method EXODUS/SLAYER.}
    \label{fig:poisson-example}
\end{figure}

Similarly as in \citet{shrestha_orchard18}, we generate a 250-dimensional input Poisson spike train across 200\,ms as well as a target spike train for a single output neuron with 4 spikes at random times. We feed it to a single hidden layer with 25 LIF neurons, which in turn connects to a single LIF output neuron. We use mean squared error loss and the ADAM optimiser, as it is invariant to different scaling factors of the gradient. We study convergence speeds for different parameters. Example output and target spike trains can be seen in Figure~\ref{fig:poisson-example} on top. The same figure also shows loss curves over 3000 epochs. Whereas EXODUS converges around epoch 1850 in this example, SLAYER fails to do so.  We repeated our experiment for different parameters, including the time constant of the membrane potential of our LIF neurons and the learning rate. We averaged the loss over 5 runs for each method with different input and target spike trains and then sum up the averaged loss. This gives us an idea of speed of convergence. Figure~\ref{fig:poisson-example} shows such summed losses for different parameters. In all cases, calculating gradients using EXODUS results in lower losses on average.  

\subsection{Gradients when output is not filtered with spike response}

\begin{align*}
    \boldx{d}{L}[n] = \deriv{\loss}{\boldx{z}{L}[n]} = \sum_{m=n}^{T} \deriv{\loss}{\boldx{s}{L}[m]} \deriv{\boldx{s}{L}[m]}{\boldx{z}{L}[n]} = \sum_{m=n}^{T} \deriv{\loss}{\boldx{s}{L}[m]} \boldx{\sigma}{L}_n[m]
\end{align*}

The term \(\boldx{e}{L}\) disappears. For \(l<L\), \(\boldx{d}{l}\) and \(\boldx{e}{l}\) are defined the same way as before.

\subsection{Training time raw numbers}
\begin{table}[h!]
    \centering
    \begin{tabular}{cccc}
        \toprule
            & \multicolumn{3}{c}{Training time per epoch [s]}\\
            & EXODUS & SLAYER & BPTT \\
        \midrule
        DVS & $\mathbf{33.4\pm0.6}$ & $42.5\pm1.8$ & $152.1\pm15.02$ \\
        HSD & $\mathbf{15.7\pm0.1}$ & $16.1 \pm 0.2 $ & $40.4 \pm 11.4 $ \\
        SSC & $\mathbf{26.0\pm0.5}$ & $68.2 \pm 0.2$ & $408.3 \pm 14.9$ \\
        \bottomrule
    \end{tabular}
    \caption{Training time is measured in seconds per epoch on a NVIDIA GeForce 1080 Ti averaged across 3 epochs.}
    \label{tab:training_time_seconds}
\end{table}

\end{document}